\documentclass[conference]{IEEEtran}
\usepackage{times}

\usepackage[numbers,square,comma,sort&compress]{natbib}
\usepackage{multicol}
\usepackage[bookmarks=true,pagebackref=true]{hyperref}
\usepackage{multirow}

\usepackage{url}
\usepackage{float}

\usepackage{comment}
\usepackage{mathtools}
\usepackage{amsfonts}  %
\usepackage{cleveref}
\usepackage{graphics}
\usepackage[pdftex]{graphicx}
\usepackage{wrapfig}
\usepackage[font=footnotesize]{caption}
\usepackage{color}
\usepackage{dsfont}
\usepackage[]{mdframed}
\usepackage{algorithm}
\usepackage{algorithmic}
\usepackage{xparse}
\usepackage{amsmath}
\usepackage{bm}
\usepackage{mathtools}
\usepackage{amssymb}
\usepackage{amsthm}
\usepackage{amssymb}

\usepackage{booktabs}
\usepackage{epigraph}
\usepackage{listings}
\usepackage{xcolor} %
\usepackage{subcaption}

\lstdefinestyle{pythonstyle}{
    language=Python,
    basicstyle=\small\ttfamily,
    keywordstyle=\color{blue},
    stringstyle=\color{red},
    commentstyle=\color{green!50!black},
    morecomment=[l]{\#},
    showstringspaces=false,
    numbers=left,
    numberstyle=\tiny\color{gray},
    frame=single,
    breaklines=true,
    tabsize=4
}

\usepackage{xcolor} %
\usepackage{hyperref} %
\hypersetup{
    colorlinks=true,
}

\usepackage{xargs} %
\usepackage[colorinlistoftodos,prependcaption,textsize=tiny]{todonotes}
\newcommandx{\wrn}[2][1=]{\todo[linecolor=red,backgroundcolor=red!25,bordercolor=red,#1]{#2}}
\newcommandx{\cmt}[2][1=]{\todo[linecolor=blue,backgroundcolor=blue!25,bordercolor=blue,#1]{#2}}

\newcommand{\ba}{\mathbf{a}}

\newcommand{\bo}{\mathbf{o}}
\newcommand{\bq}{\mathbf{q}}
\newcommand{\bI}{\mathbf{I}}
\newcommand{\bX}{\mathbf{X}}

\newcommand{\lang}{\ell}
\newcommand{\piref}{\pi_\text{ref}}
\newcommand{\rawtext}{\hat{\ell}} %

\def \Pizf {$\pi_{0.5}$}
\def \Pizs {$\pi_{0.6}$}
\def \ModelSymbol {$\pi^{*}_{0.6}$}

\def \MethodFullName {RL with Experience and Corrections via Advantage-conditioned Policies}
\def \MethodName {\textsc{Recap}}
\def \MethodSymbol {$\pi^{*}_{0.6}$}
\IEEEoverridecommandlockouts

\begin{document}
\makeatletter
\let\@oldmaketitle\@maketitle%
\renewcommand{\@maketitle}{\@oldmaketitle%
  \begin{center}
  \captionsetup{type=figure}
  \includegraphics[width=1.0\textwidth]{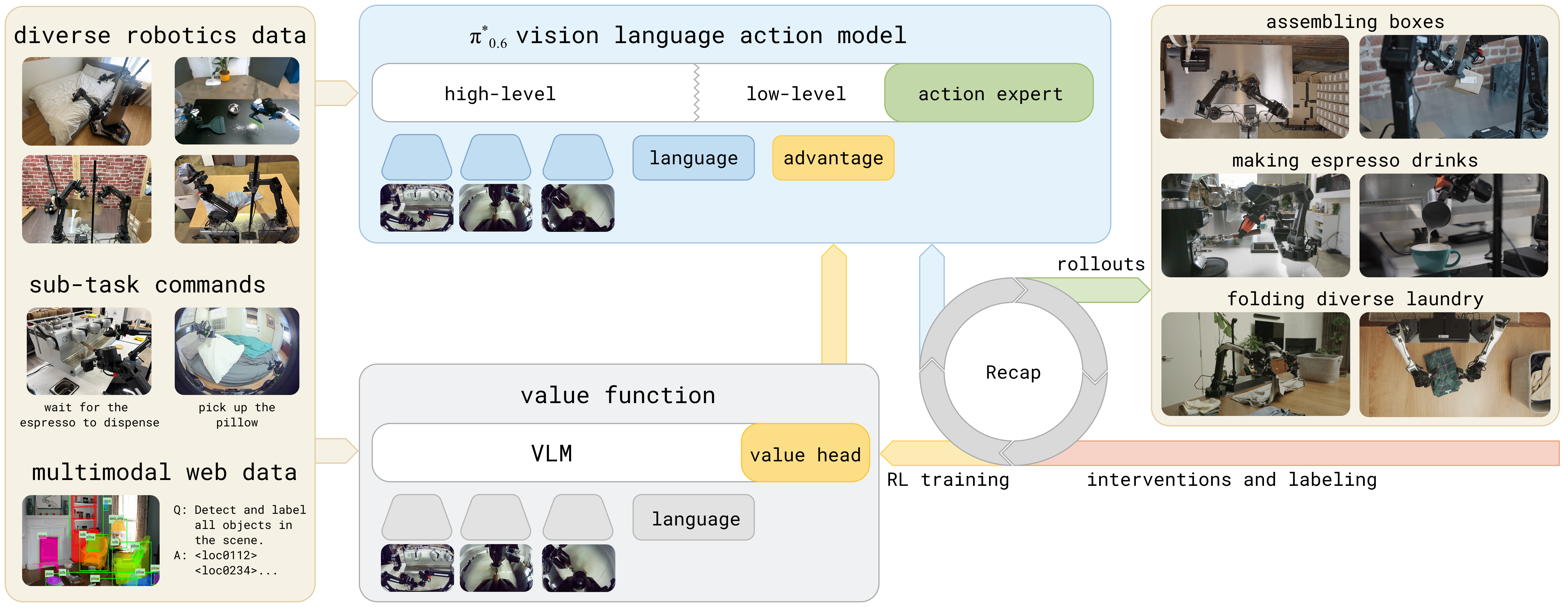}
  \caption{\textbf{\MethodName{} enables training VLAs with reward feedback and interventions.} Our system starts with a pre-trained VLA that incorporates \emph{advantage conditioning}, allowing the model to learn effectively from real-world experience. For each task, we deploy the model and collect both autonomous rollouts and online human corrections. We then fine-tune the value function on this online data, improving its estimates of how actions influence performance. Fine-tuning and conditioning the VLA on these updated advantage estimates in turn improves policy behavior.
  }
    \label{fig:teaser}
    \end{center}
}
\makeatother

\title{
\ModelSymbol: a VLA That Learns From Experience
}

\pdfinfo{
   /Author (Physical Intelligence)
   /Title  (@title)
   /Subject (Robot Foundation Models)
   /Keywords (Robot Foundation Models)
}

\author{
\centering
\begin{minipage}{0.95\textwidth}
\centering
\textbf{Physical Intelligence}\\
\vspace{0.2em}
\footnotesize
Ali Amin, Raichelle Aniceto, Ashwin Balakrishna, Kevin Black, Ken Conley, Grace Connors, James Darpinian, Karan Dhabalia, Jared DiCarlo,\\
Danny Driess, Michael Equi, Adnan Esmail, Yunhao Fang, Chelsea Finn, Catherine Glossop, Thomas Godden, Ivan Goryachev, Lachy Groom,\\
Hunter Hancock, Karol Hausman, Gashon Hussein, Brian Ichter, Szymon Jakubczak, Rowan Jen, Tim Jones, Ben Katz, Liyiming Ke, \\
Chandra Kuchi, Marinda Lamb, Devin LeBlanc, Sergey Levine, Adrian Li-Bell, Yao Lu, Vishnu Mano, Mohith Mothukuri, Suraj Nair, Karl Pertsch,\\
Allen Z. Ren, Charvi Sharma, Lucy Xiaoyang Shi, Laura Smith, Jost Tobias Springenberg, Kyle Stachowicz, Will Stoeckle, Alex Swerdlow, \\
James Tanner, Marcel Torne, Quan Vuong, Anna Walling, Haohuan Wang, Blake Williams, Sukwon Yoo, Lili Yu, Ury Zhilinsky, Zhiyuan Zhou\\[0.3em]
\normalsize
\url{https://pi.website/blog/pistar06}
\end{minipage}
}

\maketitle

\begin{abstract}
We study how vision-language-action (VLA) models can improve through real-world deployments via reinforcement learning (RL). We present a general-purpose method, \MethodFullName{} (\MethodName{}), that provides for RL training of VLAs via advantage conditioning. Our method incorporates heterogeneous data into the self-improvement process, including demonstrations, data from on-policy collection, and expert teleoperated interventions provided during autonomous execution. \MethodName{} starts by pre-training a generalist VLA with offline RL, which we call \ModelSymbol{}, that can then be specialized to attain high performance on downstream tasks through on-robot data collection. We show that the \ModelSymbol{} model trained with the full \MethodName{} method can fold laundry in real homes, reliably assemble boxes, and make espresso drinks using a professional espresso machine. On some of the hardest tasks, \MethodName{} more than doubles task throughput and roughly halves the task failure rate.
\end{abstract}

\IEEEpeerreviewmaketitle

\section{Introduction}

\setcounter{figure}{1}

\setlength{\epigraphwidth}{0.42\textwidth}
\epigraph{\textit{It’s amazing what you can learn if you’re not afraid to try.}}{Robert A. Heinlein, \textit{Have Space Suit--Will Travel}}

Practice makes perfect: while people are remarkably flexible in acquiring new skills, mastery invariably requires learning from repeated attempts. With general-purpose robotic foundation models, such as vision-language-action (VLA) models, we can flexibly specify tasks for generalist robots through prompts. But just like people, these models will need to \emph{practice} a skill to achieve mastery. This means leveraging not only on demonstration data, but also autonomously collected experiential data that allows the policy to correct the mistakes that it actually makes in deployment, improve speed and robustness beyond the level of human teleoperation, and adapt to new deployment conditions. The foundations of learning through autonomous practice, as formalized with reinforcement learning (RL)~\citep{sutton2018reinforcement}, have been known for decades, but instantiating these principles in a general and scalable robotic learning system presents significant challenges: designing scalable and stable RL methods for large models, handling heterogeneous data from different policies, and setting up RL training with reward feedback in the real world, where reward signals might be ambiguous or stochastic.

\begin{figure*}
    \centering
    \includegraphics[width=1.0\linewidth]{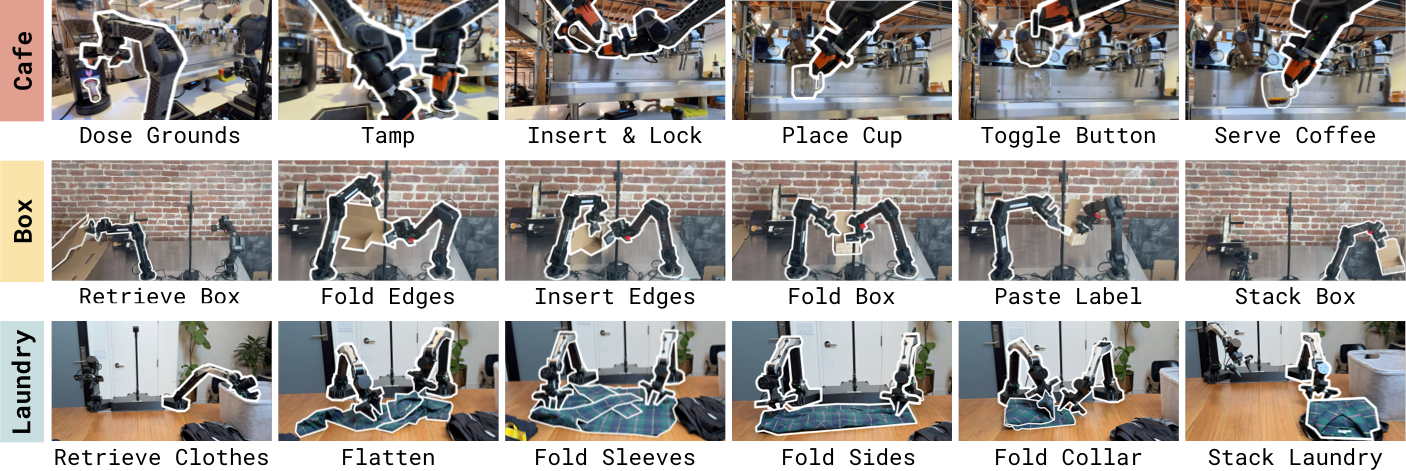}
    \caption{\textbf{Some of the tasks learned by \MethodName{}.} \ModelSymbol{} trained with \MethodName{} can make espresso drinks, assemble cardboard boxes, and fold diverse and realistic laundry with a high success rate. Each task involves realistic variability -- flattened unfolded boxes stick together and bend, making espresso drinks requires pouring liquids, and folding laundry requires generalization to a wide range of clothing items.}
    \label{fig:filmstrip}
\end{figure*}

In this paper, we present \MethodName{}, a method that enables VLA models to incorporate reward feedback in all stages of the training pipeline, from pre-training all the way to training on data from autonomous execution. \MethodName{} aims to address this problem with a general-purpose recipe that combines demonstrations, autonomous experience, and expert interventions. Starting from the training recipe for a general-purpose VLA and training on diverse data from many different robotic platforms, \MethodName{} first pre-trains the VLA with offline RL, followed by additional training on data collected through deployments. During these deployments, the robot receives (sparse) reward feedback based on the outcome of each trial, and potentially additional expert interventions that correct mistakes. The training process follows an offline RL~\citep{LangeBatchRL, levine2020offline} recipe: we train a value function that evaluates progress toward successful task completion, and then use this value function to estimate the advantage of each action in the dataset. By conditioning the policy on an improvement indicator based on this advantage~\citep{Frans2025DiffusionGuidance}, we can obtain an improved policy. Figure~\ref{fig:teaser} provides a high-level overview of \MethodName{}.

We can use \MethodName{} to train policies for complex tasks, such as folding diverse laundry, assembling boxes, or making espresso drinks. We illustrate some of these tasks in Figure~\ref{fig:filmstrip}. The method starts by pre-training the \ModelSymbol{} model with offline RL on a diverse multi-task and multi-robot dataset. \MethodSymbol{} is an adaptation of the \Pizs{} model for RL, and \Pizs{} is an improvement on \Pizf{}~\citep{black2025pi05}, adding a larger backbone and more diverse conditioning~\citep{pi06model}. \ModelSymbol{} adds the ability to condition on binarized \emph{advantage} values, which makes it possible to incorporate a value function to improve the policy. After pre-training \MethodSymbol{} finetunes the \ModelSymbol{} model to a downstream task with demonstrations, and then performs one or more iterations of on-robot data collection to improve the model with RL. Training \ModelSymbol{} with \MethodName{} on autonomous experience more than doubles the throughput on some of the hardest tasks, and can decrease failure rates by 2$\times$ or more. This enables \ModelSymbol{} to reach practically useful levels of robustness: we were able to run it to make espresso drinks for 13 hours straight, fold novel laundry items in a new home for over two hours without interruptions, and assemble boxes that are used for real packaging in a factory.

While \MethodName{} is based on individual algorithmic components that have been explored in prior works, the particular combination of these components is novel, and the results show, for the first time, that a general-purpose reinforcement learning recipe with human reward feedback and interventions can significantly improve both the robustness and throughput of VLA models with experience collected through deployment.

\section{Related Work}

Policies trained with imitation learning are known to suffer from compounding errors~\cite{ross2011dagger} and, at best, can only be as performant as the demonstration data. The goal of this work is to improve the reliability and speed of vision-language-action policies by going beyond imitation learning from offline demonstrations.
Prior works have used online interventions to improve robotic manipulation policies~\cite{Laskey2016SHIV,Laskey2017DART,jang2022bc,Hu2025RaC}. We adopt a form of such interventions, called human-gated DAgger~\cite{kelly2019hg,jang2022bc}.
In contrast to these works, our method uses both expert interventions and fully autonomous experience, resulting in an RL-based framework that integrates multiple data sources. There is a large body of work on using RL for autonomous improvement of robotic manipulation policies~\cite{levine2016end,kalashnikov2018qt,mandlekar2019iris,Sharma2023MEDALpp,Mendonca2023ALAN,Mendonca2024ContinuouslyImprovingMobileManipulation,luo2024serl,ankile2025residual,LampeMastering2024}, including methods using diffusion-based policies~\cite{Dong2025BatchOnlineIDQL,Ren2025DPPO,Lei2025RL100}, in multi-task settings~\cite{mtopt2021arxiv,Gupta2021ResetFreeMultiTask}, and using pre-trained multi-task policies~\cite{bousmalis2023robocat,Kumar2023PreTrainingForRobots,Yang2024RoboFuME}. Unlike these works, we study how to scale real-world RL to large VLA policies for long-horizon, fine-grained manipulation tasks.

Many recent works have studied how to improve a base VLA model through RL. Several works directly apply the proximal policy optimization (PPO) algorithm and variations thereof %
to VLA fine-tuning~\cite{Tan2025InteractivePostTraining,Lu2025VLARL,Liu2025WhatCanRLBringVLA,Chen2025piRL,Li2025SimpleVLA_RL}, yielding approaches that are difficult to extend to real-world RL in an efficient and scalable fashion. 
Another line of research has explored RL fine-tuning \emph{on top of} pre-trained VLA models, where RL either trains a residual policy~\citep{Guo2025ImprovingVLA,xiao2025selfimprovingvisionlanguageactionmodelsdata}, fine-tunes an action head network~\cite{chen2025conrft}, selects or refines actions proposed by the VLA~\cite{Mark2024PolicyAgnosticRL,nakamoto2025steering,zhang2025ate}, or optimizes a policy acting in the noise space of a diffusion-based VLA~\cite{Wagenmaker2025DSRL}. Some of these works have also explored ways to distill the learned behavior back into the VLA for end-to-end iterative improvement~\citep{Mark2024PolicyAgnosticRL,Guo2025ImprovingVLA,xiao2025selfimprovingvisionlanguageactionmodelsdata,xu2024rldg}. These prior works generally use discrete actions or simple Gaussian continuous action distributions. 
A critical distinction is that we train an entire VLA end-to-end using (iterated) offline RL, with an expressive flow matching VLA model. 
This is made possible by a simple and scalable advantage-conditioned policy extraction method, which removes much of the complexity of using policy gradient style objectives with large VLA models. In our comparisons, we show that this significantly outperforms a more traditional policy gradient based extraction scheme.

More closely related to \MethodName{} in terms of methodology, a number of prior works have integrated value functions and end-to-end RL training of VLAs on real robots~\citep{Huang2025CORFT,Zhang2024GRAPE,Zhai2025VLAC,Ghasemipour2025SelfImprovingEFM}. 
For example, \citet{Huang2025CORFT} apply calibrated Q-learning to an offline demonstration dataset for grasping tasks, without an online improvement phase.
\citet{Zhang2024GRAPE} use direct preference optimization (DPO) to optimize pick-and-place skills from human preferences, using online rollouts from a VLA. 
Finally,~\citet{Zhai2025VLAC,Ghasemipour2025SelfImprovingEFM} use PPO and REINFORCE respectively with time-to-completion value functions to train VLAs for tasks like moving a bowl, unfolding a mat, and pushing objects on a table.
In contrast to these prior works, we describe an iterated offline RL framework for VLAs with multiple advantages. First, our method supports high-capacity diffusion and flow-based VLAs, unlike the discrete-action models studied in prior works. Second, we avoid the need for on-policy PPO or REINFORCE by using an advantage conditioning strategy for policy extraction, which can utilize all prior (off-policy or offline) data. Lastly, our evaluation consists of complex, dexterous, and temporally extended tasks, where our method increases throughput by about 2$\times$ while handling deformable objects, liquids, and multi-stage tasks.

Prior works have explored the idea of conditioning the policy on rewards, values, and advantages~\cite{Schmidhuber2019UpsideDownRL,Kumar2019RewardConditionedPolicies,Chen2021DecisionTransformer,Brandfonbrener2022RCSL,Emmons2022RvS,Furuta2022GDT,Yamagata2023QDT,Zheng2022ODT,kuba2023advantage,Wu2023ElasticDecisionTransformer}, including methods that use classifier-free guidance~\cite{Frans2025DiffusionGuidance}. We extend this approach to pre-train and fine-tune a large-scale generalist VLA policy~\cite{black2025pi05}, incorporating a variety of data sources (including demonstrations, interventions, and autonomous policy roll-outs) to learn real robotic manipulation tasks.
Recent research has also studied how to effectively train multi-task, language-conditioned reward functions~\cite{Shao2020Concept2Robot,Chen2021DVD,Nair2022LanguageConditionedRobotBehavior,Sontakke2023RoboCLIP,Yu2023LanguageToRewards,Zhang2025ReWiND,Alakuijala2025VideoLanguageCritic} and value functions~\cite{Ma2023LIV,Ma2025GVL,Zhai2025VLAC}. Building on these works, we also train a language-conditioned distributional value function, which allows us to estimate state-action advantages for our advantage-conditioned VLA training framework.

\section{Preliminaries}
\label{sect:prelim}

\noindent \textbf{Reinforcement learning.} We consider the standard RL setting in which an agent, given by a policy $\pi(\ba_t | \bo_t)$, selects actions $\ba_t$ given an observation $\bo_t \in \mathcal{O}$. 
We define a trajectory as \mbox{$\tau = (\bo_0, \ba_0, \cdots, \bo_T) \in \mathcal{O} \times \mathcal{A} \cdots \mathcal{O}$}. A distribution over trajectories $\rho_\pi(\tau)$ is induced by the policy $\pi(\ba_t | \bo_t)$ and the stochastic dynamics $p(\bo_{t+1} | \bo_t, \ba_t)$: \mbox{$\rho_{\pi}(\tau) = p(\bo_0) \prod_{t=0}^{T-1} \pi(\ba_t | \bo_t) p(\bo_{t+1} | \bo_t, \ba_t)$}.\footnote{For simplicity, we assume the observation $\bo_t$ constitutes a valid Markovian state. While not true in general, it is a common simplification in robotic RL.} The reward function is given by $r(\bo_t, \ba_t)$, and we abbreviate it to $r_t$ to shorten notation, where $r_T$ is the terminal reward. We can define the discounted cumulative reward, or return, as $R(\tau) = \sum_{t=0}^T r_t$ (we do not use a discount factor, though one could easily be added). The goal of RL is to maximize the cumulative reward (or return), learning a policy that maximizes $\mathcal{J}(\pi) = \mathbb{E}_{\tau \sim \rho_{\pi}}[R(\tau)] = \mathbb{E}_{\tau \sim \rho_{\pi}}[\sum_{t=0}^T r_t]$. The value function for a policy $\pi$ is then defined as $V^\pi(\bo_t) = \mathbb{E}_{\tau_{t+1:T}} [\sum_{t=t}^T r_t]$. We can then calculate an advantage value for an action $\ba_t$ as $A^\pi(\bo_t,\ba_t) = \mathbb{E}_{\rho_{\pi}(\tau)} [\sum_{t'=t}^{t+N-1} r_{t'} + V^\pi(\bo_{t+N})] - V^\pi(\bo_t)$, corresponding to an n-step estimate.

\noindent \textbf{Regularized reinforcement learning.} Instead of maximizing $\mathcal{J}(\pi)$, it is common to use regularization in RL, optimizing for a policy that maximizes reward while remaining close to some reference policy $\piref$~\citep{schulman2017proximal,abdolmaleki2018maximum,peng2019advantage,Dayan1997UsingEM,PetersREPS}. This is important, for example, when we want to train for many gradient steps on the same data, in which case $\piref$ typically corresponds to the behavior policy that collected the training data. This can be formalized via the objective 
$\mathcal{J}(\pi, \piref) = \mathbb{E}_{\tau \sim \rho_{\pi_\theta}}[\sum_{t=0}^T \gamma^t r_t] - \beta \mathbb{E}_{\bo \sim \rho_{\pi_\theta}} [D(\pi(\cdot | \bo) \| \pi_\text{ref}(\cdot | \bo)) ], $ where $D$ denotes some divergence metric. For the case where $D$ is the KL divergence, we have the well-known result that \mbox{$\hat{\pi}(\ba | \bo) \propto \piref(\ba | \bo) \exp(A^{\piref}(\bo, \ba) / \beta)$} is the solution to $\max_\pi J(\pi, \piref)$, with Lagrange multiplier $\beta$~\citep{abdolmaleki2018maximum,peng2019advantage,Dayan1997UsingEM,PetersREPS}. Our advantage-conditioned policy extraction method is based on a closely related but less well-known result: if we define the policy \mbox{$\hat{\pi}(\ba | \bo) \propto \piref(\ba | \bo) p(I | A^{\piref}(\bo, \ba))^\beta$},
where $p(I | A^{\piref}(\bo, \ba)) = g(A^{\piref}(\bo, \ba)) / \int g(A^{\piref}(\bo, \ba')) \mathrm{d}\ba'$ is the probability of any action $\ba$ improving over $\piref$ as measured by a monotonically increasing function $g$, then $\hat{\pi}$ is guaranteed to improve over $\piref$, i.e., $\mathcal{J}(\hat{\pi}) \geq \mathcal{J}(\piref)$ \citep{wangmarwil,Frans2025DiffusionGuidance}. We will use this property in deriving our policy extraction method in Section~\ref{sect:adv_cond}.
Using this definition we can then obtain a parametric policy from the closed form definition of $\hat{\pi}$ by solving the following minimization problem: $\min_\theta \mathbb{E}_{s \sim \rho_{\pi_{\text{ref}}}} [KL(\hat{\pi}, \pi_\theta)]$.

\section{\MethodFullName{} (\MethodName{})}

\begin{figure}[t]
    \centering
    \includegraphics[width=\linewidth]{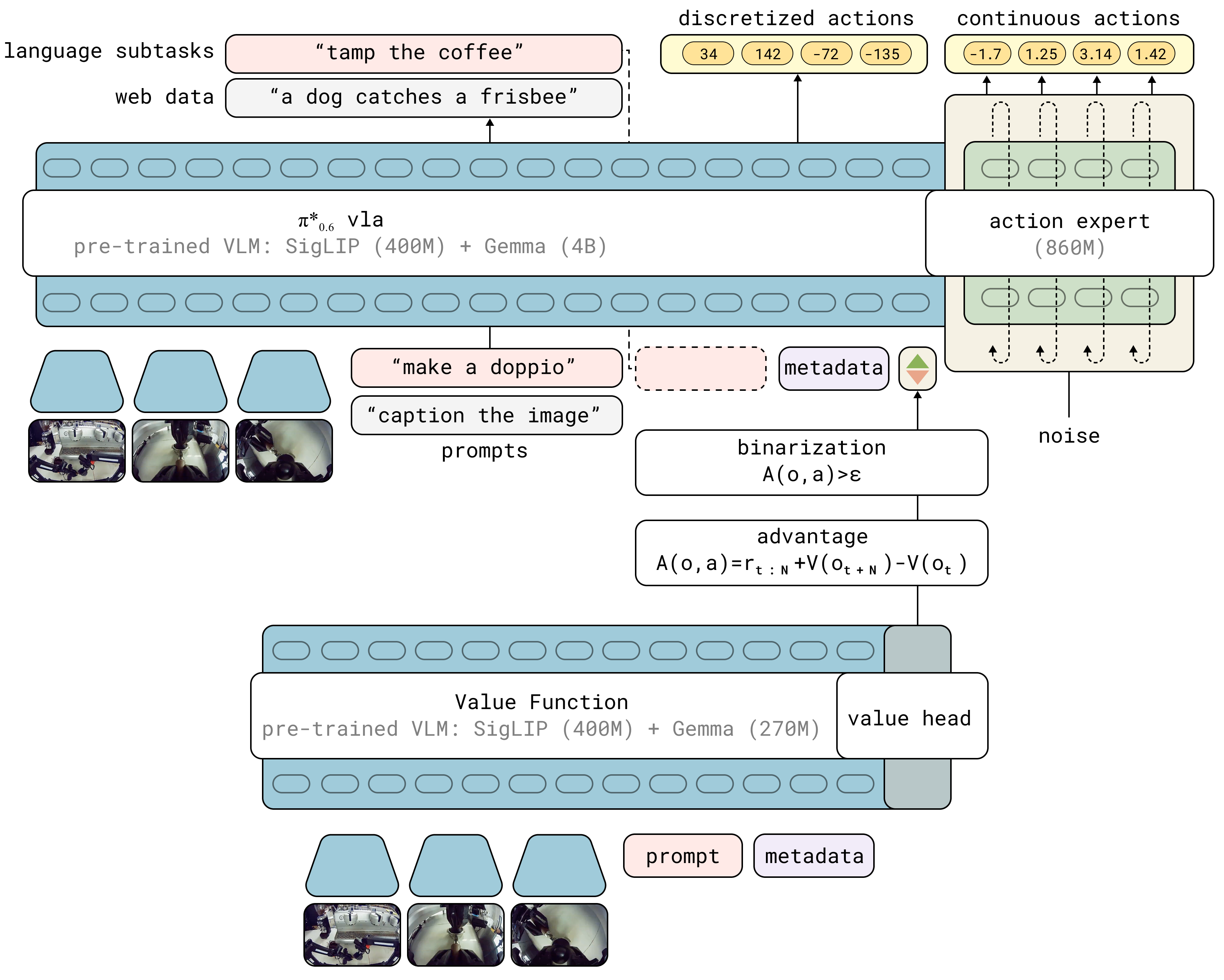}
    \caption{\textbf{Interaction between the \ModelSymbol{} VLA and value function during \MethodName{} training.} The \ModelSymbol{} VLA uses a pre-trained VLM backbone. Training follows the KI recipe~\citep{driess2025knowledge}, with next-token prediction on many data sources in pre-training, and an flow-matching action-expert with stop gradient. The VLA is conditioned on a binarized advantage indicator, obtained from a separate value function initialized from a pre-trained but smaller VLM model.}
    \label{fig:arch}
\end{figure}

Our method consists of the follow steps, which can be repeated one or more times to improve a base VLA model:
\begin{enumerate}
    \item \textbf{Data collection.} We run the VLA on the task, labeling each episode with task outcome labels (which determine the reward), and optionally providing human interventions to provide examples of corrections for mistakes in the earlier iterations.
    \item \textbf{Value function training.} We use all of the data collected so far to train a large, multi-task value function, which we refer to as $V^{\piref}$, that can detect failures and judge the expected time to task completion.
    \item \textbf{Advantage conditioned training.} To improve the VLA policy with this value function, we include an optimality indicator based on advantage values derived from this value function in the VLA prefix. This ``advantage conditioned'' recipe provides a simple and effective way to extract a more optimal policy from our value function with suboptimal data.
\end{enumerate}
Figure~\ref{fig:teaser} illustrates the overall structure of the training process, while Figure~\ref{fig:arch} provides more detailed specifics of the value function and policy architectures.
Our pre-training phase consists of performing steps (2) and (3) above on our entire pre-training dataset, which consists of tens of thousands of hours of demonstrations from numerous tasks and a variety of different robots. Then, we perform steps (1), (2), and (3) one or more times to further improve the VLA with autonomously collected data.
We describe the value function training and policy training steps below, and then present our specific instantiation of this approach for training \ModelSymbol{} in Section \ref{sect:system}.

\subsection{Distributional value function training}
\label{sect:vf_train}

To train a value function that can act as a reliable critic for any task in our pre-training or post-training stages, we represent $V^{\piref}$ with a multi-task distributional value function \mbox{$p_\phi(V | \bo_t, \lang) \in \Delta_B$} \citep{bellemare2017distributional}, mapping the observations $\bo_t$ and language command $\lang$ to a distribution over $B$ discretized value bins. In our implementation, this value function uses the same architecture as the VLA policy, but with a smaller VLM backbone. Using $R_t(\tau) = \sum_{t' = t}^T r_{t'}$ to denote the empirical return of a trajectory $\tau$ from time step $t$ until the end, we train $p_\phi(V | \bo_t, \lang)$ by first discretizing the empirical return value $R_t(\tau)$ into $B = 201$ bins (using $R^B_t$ to denote the discretized returns), and then minimizing the cross-entropy $H$ over the trajectories in the current dataset $\mathcal{D}$:
\begin{equation}
    \min_\phi \mathbb{E}_{\tau \in \mathcal{D}} \left[ \sum_{\bo_t \in \tau} H(R^B_t(\tau), p_\phi(V | \bo_t, \lang) ) \right].
    \label{eq:value_function_objective}
\end{equation}
This is a Monte Carlo estimator for the value function of the policy represented by the dataset $\mathcal{D}$ (i.e., the behavior policy $\piref$). We can extract a continuous value function (and thus an advantage) from the learned value distribution using $V^{\piref}(o_t, \lang) = \sum_{b\in [0, B]}p_\phi(V=b | \bo_t) v(b),$ where $v(b)$ denotes the value corresponding to bin $b$. During the pre-training phase, the dataset $\mathcal{D}$ corresponds to the human demonstrations, and the value function captures the expected return for the task and metadata we condition on, while on subsequent iterations, it skews toward a weighted combination of the return of the demonstrations and the learned policy.

While this on-policy estimator is less optimal than a more classic off-policy Q-function estimator, we found it to be simple and highly reliable, while still allowing for substantial improvement over imitation learning. Our method could be extended to accommodate off-policy estimators in future work.

\begin{figure*}[t]
    \centering
    \includegraphics[width=0.99\linewidth]{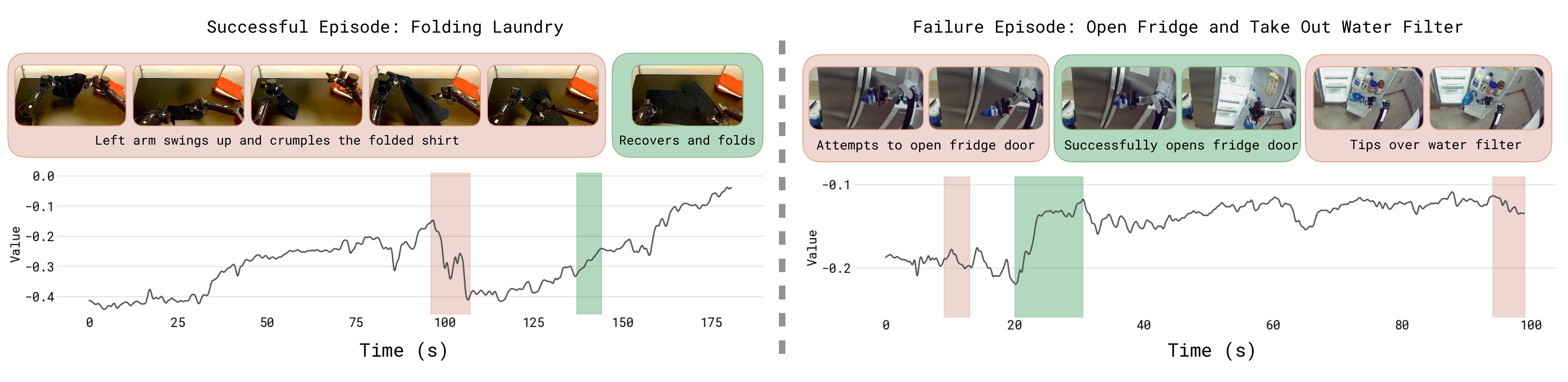}
    \caption{\textbf{Visualization of the value functions.} We train a multi-task value function to predict the number of steps to success, normalized by maximum task length to $(-1, 0)$, where $0$ corresponds to successful completion.
    We visualize the value function output on a folding task that finished successfully (left), and an unsuccessful example of a manipulation task from the pre-training dataset (right).
    The red parts highlight a drop in value, and green parts highlight increases; images on top show the corresponding frames of the episode.
    The visualization shows that the VF correctly identifies mistakes in the episode, as well as the speed of progress.
    }
    \label{fig:vf-viz}
\end{figure*}

\subsection{Policy extraction via advantage conditioning}
\label{sect:adv_cond}

Once we have the value function $V^{\piref}$, we need a way to train an improved policy using this value function. This is called \emph{policy extraction}. An effective policy extraction method in our setting needs to satisfy several criteria. First, it needs to effectively utilize diverse off-policy data, comprising the initial demonstrations, the expert interventions, and autonomous episodes from both the latest policy and older policies. This is closely related to the challenge faced by offline RL methods~\citep{LangeBatchRL,levine2020offline}. Second, it needs to be scalable and easily to apply to large VLA models, including models that use flow matching or diffusion to generate actions. Third, it needs to effectively utilize both good (near-optimal) and bad (suboptimal) data, which is important if we want to improve the policy using autonomous experience.

Among the existing methods for policy extraction, policy gradient methods (including regularized policy gradients and reparameterized gradients) are perhaps the most widely used~\citep{schulman2017proximal, haarnoja2018sac}, but these methods are difficult to apply to flow matching models, which do not readily provide a tractable log-likelihood, making them hard to scale up to modern VLA architectures (see comparisons in Section~\ref{sec:experiments}). An alternative is to use weighted regression methods, such as AWR~\citep{peng2019advantage,wangCRR,kostrikov2022offline}, which implicitly provide for regularization to the behavior policy and use a simple (importance-weighted) supervised learning objective. However, these methods discard or significantly downweight a significant portion of the data, effectively implementing a kind of filtered imitation technique. Instead, we use a variant of \emph{advantage conditioning}~\citep{Kumar2019RewardConditionedPolicies}, where the policy is trained on all of the data with supervised learning, but with an additional input indicating how \emph{optimal} the action is based on the advantage. This is closely related to a variety of methods in the literature that propose to condition the policy on some function of the resulting trajectory~\citep{Schmidhuber2019UpsideDownRL,Brandfonbrener2022RCSL}.

The specific formulation in our method is most closely related to CFGRL~\citep{Frans2025DiffusionGuidance}. Building on the formulation in Section~\ref{sect:prelim}, we can apply Bayes rule to rewrite the probability of policy improvement  as $p(I | A^{\piref}(\bo, \ba)) = \pi_{\text{ref}}(\ba | I, \bo) / \pi_{\text{ref}}(\ba | \bo)$. Applying this to our setting and including language conditioning, we can obtain an alternative closed form for the improved regularized policy described in Section~\ref{sect:prelim} as
\begin{equation}
      \hat{\pi}(\ba,  | \bo, \lang) \propto \piref(\ba | \bo, \lang) \left( \frac{\piref(\ba | I, \bo, \lang)}{\piref(\ba | \bo, \lang)} \right)^\beta.
\label{eq:improved_policy}
\end{equation}
For the special case $\beta = 1$, $\hat{\pi}(\ba,  | \bo, \lang) = \piref(\ba | I, \bo, \lang)$.

We can therefore represent $\hat{\pi}$ without needing to explicitly represent the improvement probability $p(I | A^{\piref}(\bo, \ba))$, if we train the policy so that it can represent both $\piref(\ba | \bo, \lang)$ and $\piref(\ba | I, \bo, \lang)$. This principle is similar to the approach in classifier-free guidance, where a diffusion model is trained to model the data both with and without a conditioning variable~\citep{Frans2025DiffusionGuidance}.
We assume the improvement indicator $I$ follows a delta distribution $$
p(I | A^{\piref}(o, a, \lang)) = \delta (A^{\piref}(o, a, \lang) > \epsilon_\lang),
$$ with a task dependent improvement threshold $\epsilon_\lang$. This threshold allows us to control the optimality indicator, and minimizes the need for finding an attenuation factor $\beta$ to sharpen the improvement conditioned distribution after training.\footnote{Prior work \citep{Frans2025DiffusionGuidance} instead uniformly chose $\epsilon = 0$ and tuned $\beta$ at test time, as in classifier-free guidance (CFG). However, high CFG weights can drive the action distribution to the corners of its support (leading to aggressive behavior) and would not affect the autoregressive part of the model. We found it easier to obtain good results by instead using the threshold $\epsilon_\lang$ to trade off regularization and optimality.} The policy objective then corresponds to minimizing the following negative log-likelihood:
\begin{equation}
\begin{aligned}
\min_\theta \: &\mathbb{E}_{\mathcal{D}_{\piref}} \Big[ -\log \pi_\theta(\ba_t | \bo_t, \lang) - \alpha  \log \pi_\theta(\ba_t | I_t, \bo_t, \lang)\Big], \\
 & \text{where } I_t = \mathds{1}\big(A^{\piref}(\bo_t, \ba_t, \lang) > \epsilon_\lang \big).
\end{aligned}
\label{eq:obj_adv_cond}
\end{equation}
The advantage values $A^{\piref}(\bo_t, \ba_t, \lang)$ are obtained from the value function in the previous section, and $\alpha$ is a trade-off hyperparameter. In practice, the dataset $\mathcal{D}_{\piref}$ consists of all of the data collected so far, including all demonstrations and autonomous task attempts, and the reference policy $\piref$ is therefore a mixture of human behavior and previously deployed policies. To include human corrections, we found it useful to force $I_t = \text{True}$ (i.e., positive) for actions provided as human corrections during autonomous rollouts. This choice is reasonable if we assume that human experts always provide good corrective actions. As we will discuss in Section~\ref{sect:system}, in practice our VLA model produces both discrete and continuous outputs, with the continuous distribution represented via flow matching. Therefore, the real training objective combines likelihoods for the discrete values with the flow matching objective for the continuous values.

In practice, we pre-train one model to represent $\pi_\theta(\ba_t | I_t, \bo_t, \lang)$ on our entire pre-training dataset, and then perform one or more iterations of our method with on-policy rollouts (and, optionally, expert corrective interventions) for each task.

\subsection{Method summary}
\label{sec:summary}

We provide an overview of our full method in Algorithm~\ref{alg:summary}. As summarized at the beginning of this section, the method can be fully defined through application of three subroutines: collecting data through autonomous rollouts (with optional corrective interventions from an expert), training a value function according to Equation~\ref{eq:value_function_objective}, and training a policy according to Equation~\ref{eq:obj_adv_cond}. The only thing that changes between different steps of the method is the data provided to each subroutine: the pre-training stage uses all prior demonstration data, and the training process for the specialists for each skill $\lang^{(i)}$ uses additional autonomous data. In practice, the specialists are fine-tuned from the pre-trained model, while the final generalist is trained from scratch. Additional details on the method are provided in Appendix~\ref{app:alg_details}.

\begin{algorithm}[h]
\caption{\MethodFullName{} (\MethodName{})}
\label{alg:summary}
\begin{algorithmic}[1]
\REQUIRE multi-task demonstration dataset $\mathcal{D}_\mathrm{demo}$

\STATE Train $V_\mathrm{pre}$ on $\mathcal{D}_\mathrm{demo}$ using Eq.~\ref{eq:value_function_objective}
\STATE Train $\pi_\mathrm{pre}$ on $\mathcal{D}_\mathrm{demo}$ using Eq.~\ref{eq:obj_adv_cond} and $V_\mathrm{pre}$

\STATE Initialize $\mathcal{D}_{\lang}$ with demonstrations for $\lang$
\STATE Train $V_{\lang}^{0}$ from $V_\mathrm{pre}$ on $\mathcal{D}_{\lang}$ using Eq.~\ref{eq:value_function_objective}
\STATE Train $\pi_{\lang}^{0}$ from $\pi_\mathrm{pre}$ on $\mathcal{D}_{\lang}$ using Eq.~\ref{eq:obj_adv_cond} and $V_{\lang}^{0}$\!\!
\FOR{$k = 1$ to $K$}
    \STATE Collect data with $\pi_{\lang}^{k-1}$, add it to $\mathcal{D}_{\lang}$
    \STATE Train $V_{\lang}^{k}$ from $V_\mathrm{pre}$ on $\mathcal{D}_{\lang}$ using Eq.~\ref{eq:value_function_objective}
    \STATE Train $\pi_{\lang}^{k}$ from $\pi_\mathrm{pre}$ on $\mathcal{D}_{\lang}$ using Eq.~\ref{eq:obj_adv_cond} and $V_{\lang}^{k}$
\ENDFOR

\end{algorithmic}
\end{algorithm}

\section{Implementation, Model, and System Details}
\label{sect:system}

We instantiate \MethodName{} with a VLA that we call \ModelSymbol{}. \ModelSymbol{} is based on the \Pizs{} VLA, which is an evolution of the \Pizf{} VLA~\citep{black2025pi05} with a few improvements that we detail in the accompanying model card~\citep{pi06model}. \ModelSymbol{} additionally adds the ability condition on the binarized advantage indicator $I_t$, making it suitable for RL training with \MethodName{}. The model architecture is illustrated in Figure~\ref{fig:arch}. We train a value function alongside the VLA, following the method described in Section~\ref{sect:vf_train}. This value function is also initialized from a VLM. Training this value function and VLA with \MethodName{} results in our final model, which we call \ModelSymbol{}. In this section, we first elaborate on the design of our model and how it can be extended to use advantage values from the value function, then describe the reward function and value function, and then elaborate on the training and data collection process in our implementation.

\subsection{The \Pizs{} model}
\label{sect:pi06}

The \Pizs{} model~\citep{pi06model} is derived from the \Pizf{} model, which can flexibly represent chunked action distributions via flow matching and produce intermediate text for high-level policy reasoning. It uses the Knowledge Insulation (KI) training procedure~\citep{driess2025knowledge}, which trains the entire model end-to-end on continuous actions and discretized tokens (including actions discretized via FAST~\citep{pertsch2025fast}), while using a stop gradient to prevent the flow-matching action expert from impacting the rest of the model. Pre-training uses both robot data and vision-language co-training data from the web.

\Pizs{} improves on \Pizf{} in several ways: (i) The pre-training dataset is augmented with additional data from multiple robot platforms. (ii) The base VLM is Gemma 3~\citep{gemmateam2025gemma3technicalreport} 4B model. (iii) The size of the action expert is increased to 860M parameters.

The model can be written as $\pi_\theta(\ba_{t:t+H}, \rawtext \vert \bo_t, \lang)$, where $\bo_t = [\bX^1_t, ..., \bX^n_t, \bq_t]$
contains camera images $\bX$, the robot's configuration $\bq$, and $\lang = \lang_t + s$ is the language input consisting of the overall task prompt $\lang_t$ (e.g., ``make me an espresso''), as well as additional language inputs $s$ providing metadata that further modulates how the task is performed. 
The model produces action chunks $\ba_{t:t+H}$, which consists of joint angles and gripper commands at 50 Hz, using a separate ``action expert'' --- a dedicated set of weights (860M parameters) that are trained with flow matching specifically for action generation, but can attend to the activations in the rest of the model. The model also produces tokenized discrete outputs $\rawtext$, which includes a textual representation of the next predicted sub-task (such as ``pick up the coffee cup'') used for high-level decision-making. Since the actions are generated after $\rawtext$, action generation is effectively conditioned on this predicted sub-task, providing high-level guidance. At inference time, the sub-task prediction runs at a lower frequency than action generation. During training, the model also predicts a tokenized representation of the action chunk $\ba_{t:t+H}$, using the FAST tokenizer~\citep{pertsch2025fast}, as part of the KI recipe~\citep{driess2025knowledge}. We denote these discretized actions $a^{\ell}_{t:t+H}$. The action expert does not receive these as input, such that discrete and continuous actions are predicted independently. This results in the final training log-likelihood $\log \pi_\theta(\ba_{t:t+H}, a^{\ell}_{t:t+H}, \rawtext \vert \bo_t, \lang)$. Since we predict $\rawtext$ first, we can factorize this log-likelihood according to:
\begin{align*}
\log \pi_\theta\big(&\,\ba_{t:t+H}, a^{\ell}_{t:t+H}, \rawtext \vert \bo_t, \lang\big) = 
\log \pi_\theta\big(\rawtext \vert \bo_t, \lang\big) \\
& + \log \pi_\theta\big(a^{\ell}_{t:t+H} \vert \bo_t, \lang, \rawtext\big) + \log \pi_\theta\big(\ba_{t:t+H} \vert \bo_t, \lang, \rawtext\big).
\end{align*}

\subsection{From \Pizs{} to \ModelSymbol{} with advantage conditioning}

To incorporate information about the advantage into the policy, we expand the model inputs to contain an additional improvement indicator as an additional text input, inputting ``Advantage: positive'' when $I_t = \text{True}$, and ``Advantage: negative'' otherwise.
The VLA model is otherwise the same as described in Section~\ref{sect:pi06}. The advantage indicator appears in the training sequence after $\rawtext$ but before the (discretized and continuous) actions, such that only the action log-likelihoods are affected. The continuous part of the log-likelihood cannot be evaluated exactly, and instead is trained via the flow matching loss~\citep{lipman2022flow}. It is possible to draw a close parallel between flow matching and diffusion (under some assumptions), and the latter in turn can be interpreted as a lower bound on the log-likelihood~\citep{KingaDiffELBO}, so we can roughly motivate the sum of the log-likelihood of the discrete actions and the flow matching loss on the continuous actions as a lower bound on the overall action likelihood:
\begin{equation}
\begin{aligned}
    \log \pi_\theta(&\ba_{t:t+H}, a^{\ell}_{t:t+H} \vert I_t, \bo_t, \lang, \rawtext) \geq \\  \mathbb{E}_{\eta, \omega} \Big[& \log p_\theta(a^{\ell}_{t:t+H} |  I_t, \bo_t, \lang, \rawtext) -
     \\ & \alpha_\eta \left\|\omega - \ba_{t:t+H} - f_\theta(\ba^{\eta, \omega}_{t:t+H}, I_t, \bo_t, \lang, \rawtext)\right\|^2 \Big] 
\end{aligned}
\label{eq:cotraining},
\end{equation}
with $\ba_{t:t+H}^{\eta, \omega} = \eta \ba_{t:t+H} + (1-\eta)\omega$, $\omega \sim \mathcal{N}(0,\bI)$ denoting the noised action, where $\eta \in [0,1]$ is the flow matching time index and $f_\theta$ denotes the continuous outputs of the diffusion expert. $\alpha_\eta$ is a loss weighting term (which can optionally be noise dependent). Full details for the loss are provided in Appendix~\ref{sect:lower_bound_loss}.

During training, we randomly omit the indicator $I_t$ instead of tuning the loss multiplier $\alpha$ to allow us to either directly sample from the policy with $I_t = \text{True}$ (which corresponds to setting $\beta = 1$ in Equation~\eqref{eq:improved_policy}), or to use both a conditional and unconditional model to implement classifier-free guidance (CFG), which enables inference with $\beta > 1$. See Appendix~\ref{sect:cfg} for details.

\subsection{Reward definition and value function training}

Since our aim is to develop a general and broadly applicable method for training VLAs from experience, we use a general sparse reward definition that can be applied to essentially any task. For each episode, we obtain a label indicating whether that episode was successful. We derive the reward from this episode-level success label such that the value function corresponds to the (negative) number of steps until successful completion of the episode. This is equivalent to the following reward function, where $T$ corresponds to the last step in the episode, and $C_\text{fail}$ is a large constant that is chosen so as to ensure that failed episodes have low values:
\begin{equation}
r_t = \begin{cases}
0 &\ \text{if t = T and success} \\
-C_\text{fail} &\ \text{if t = T and failure} \\
-1 &\ \text{otherwise}.
\end{cases}
\end{equation}

With this reward function, we train the value function to predict the (negative of the) number of remaining steps until success for successful episodes, and a large negative value for failed episodes. In practice, we normalize the values predicted to be between $(-1, 0)$. Since we train on diverse tasks that have very different typical lengths, we normalize the values per task based on the maximum episode length of the task.

The value function takes as input the same language inputs as the \ModelSymbol{} VLA, and uses the same architecture design, with a smaller 670M parameter VLM backbone that is also initialized from Gemma 3 (see Figure~\ref{fig:arch}). To prevent overfitting, we also co-train the value function on a small mixture of multi-modal web data. Figure~\ref{fig:vf-viz} show visualizations of the value function on some examples of successful and failure episodes, with additional visualizations in Figure~\ref{fig:app-vf-viz} in Appendix~\ref{app:value}.

\subsection{Pre-training, data collection, and learning from experience}
\label{sec:training}

The data mixture used in the pre-training phase of our model largely follows the recipe used by \Pizf~\citep{black2025pi05}, with vision-language data from the web, prediction of subtasks $\rawtext$, and prediction of low-level actions on a variety of tasks from many different robots. We note that, after pre-training, \ModelSymbol{} can perform many more tasks than the ones used in evaluation in Section~\ref{sec:experiments}. During pre-training, we first train the value function on the same dataset, predicting (the negative of) the number of steps to successful completion of each task. Then we estimate the per-task improvement threshold, $\epsilon_\lang$, used in determining the advantage-based improvement indicator $I_t$. We set $\epsilon_\lang$ to the $30\%$ percentile of values predicted by the value function for the task $\lang$. We then run the value function on-the-fly during VLA training to estimate $A^{\piref}(\bo_t, \ba_t, \lang)$ for each example, and then use it to compute $I_t$ based on $\epsilon_\lang$. $I_t$ is included as an input to \ModelSymbol{} as described in Section~\ref{sect:pi06}. As we use a relatively small VLM backbone (670M) for the value function, on-the-fly inference of the value function incurs minimal additional cost during VLA training.

After pre-training we start a policy improvement loop for the target task. We first finetune \ModelSymbol{} with demonstration data $\mathcal{D}_{\lang}$ for the target task $\lang$. We fix the indicator $I_t$ to $\text{True}$ in this stage, which we found to lead to slightly better results, such that this stage corresponds to supervised finetuning (SFT). This results in the initial policy $\pi^0_{\lang}$, which is then used to collect additional data that is added to $\mathcal{D}_{\lang}$. While some of the episodes are collected fully autonomously, some are monitored by an expert teleoperator who can intervene to provide corrections. These corrections can show the policy how to avoid catastrophic failures or how to recover from mistakes. Note, however, that the corrections alone are unlikely to fix all issues: intervening during autonomous execution is a disruptive event, and even expert human operators cannot guarantee a consistent quality of interventions nor improve subtle aspects of the behavior, such as overall speed. Thus, the corrections serve more to fix large mistakes and overcome challenges with exploration, and do not by themselves provide for optimal supervision, in contrast to theory~\citep{ross2011dagger}. Recall from Section~\ref{sect:adv_cond} that we force $I_t = \text{True}$ for all corrections, but otherwise the entire episode (both the autonomous parts and the corrections) are optionally added to the dataset $\mathcal{D}_{\lang}$ regardless of whether or not a correction was provided.

After data collection, we finetune the value function on all of the data collected for the task so far, and then use it to finetune the policy with updated indicators $I_t$, using the same procedure as in pre-training. Both the value function and policy are finetuned from the pre-trained checkpoint, rather than the policy and value function from the last iteration. We found this to be useful for avoiding drift over multiple iterations, though it may be possible to also obtain good results by consistently finetuning from the last model.

We can repeat this process for several iterations as needed, though in practice we found that even one iteration often leads to significantly improved results.

\begin{figure}
    \centering
    \includegraphics[width=0.9\linewidth]{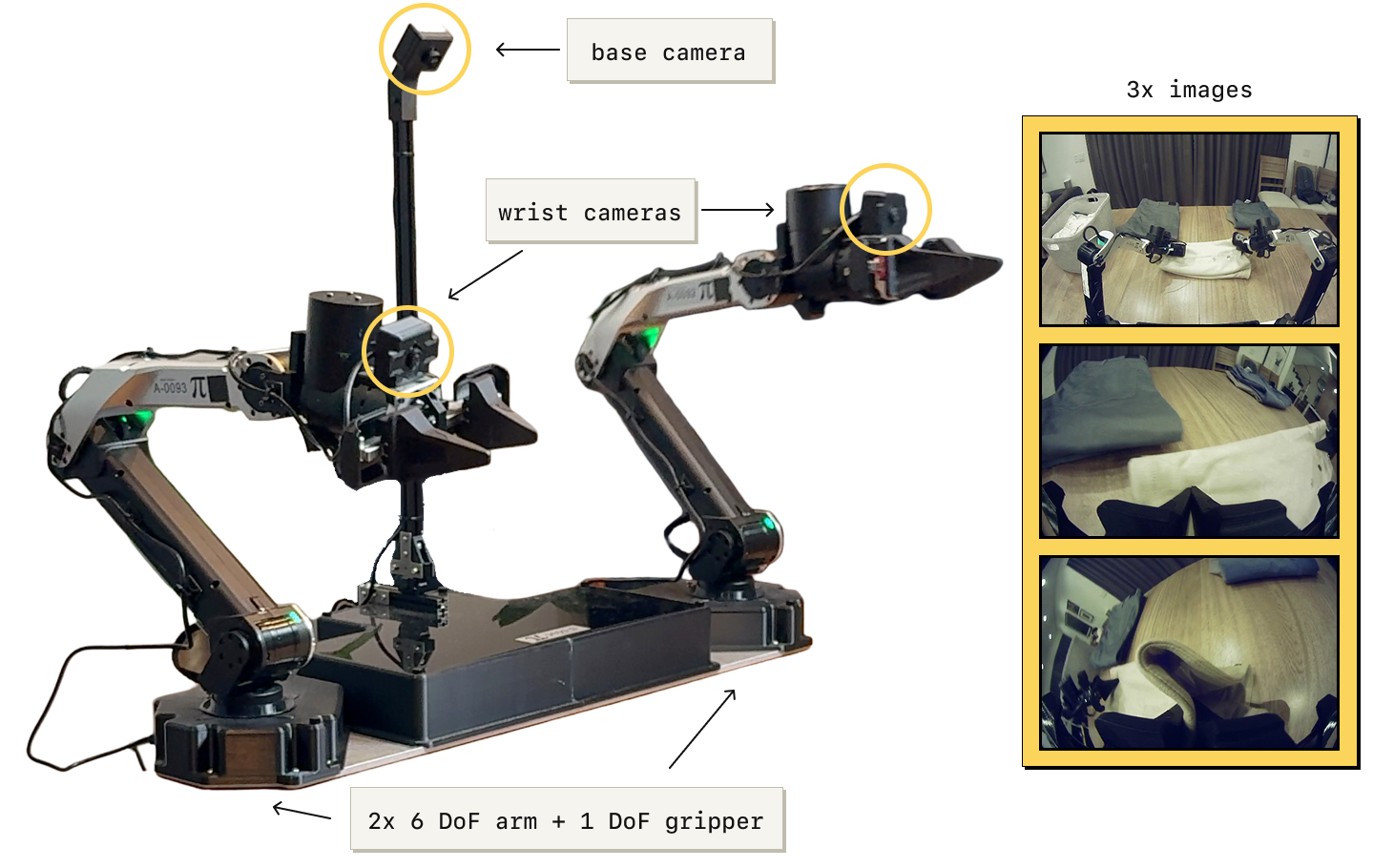}
    \caption{\textbf{The robot setup used in our experiments.} \ModelSymbol{} is trained on data from many different robots in pre-training. For the iterative improvement experiments, we use a static bimanual system with two 6 DoF arms with parallel jaw grippers. The arms are controlled at 50 Hz with joint positions. Observations consist of joint and gripper positions, as well as images from three cameras: a base camera mounted between the arms, and a wrist-mounted camera on each arm. The setup can be mounted flexibly, e.g.\ on a table.}
    \label{fig:robot}
\end{figure}

\section{Experimental Evaluation}
\label{sec:experiments}

\begin{figure*}
    \centering
    \includegraphics[width=0.95\linewidth]{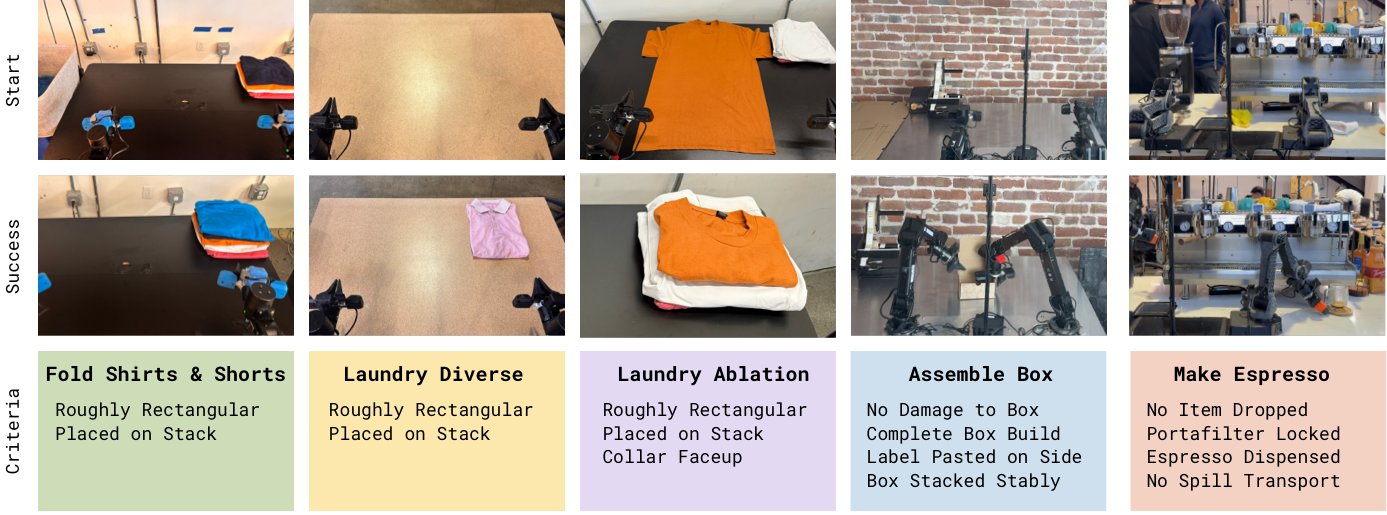}
    \caption{\textbf{Illustrations of the tasks used in our experiments.} Tasks include three different laundry variants, assembling boxes, and making coffee drinks with an espresso machine.}
    \label{fig:tasks}
\end{figure*}

In our experimental evaluation, we use \MethodName{} to train the \Pizs{} model on a set of realistic tasks: making espresso drinks, folding diverse laundry, and assembling boxes. Each task requires multiple steps, ranging from 5 to 15 minutes in duration, complex manipulation behaviors (constrained forceful manipulation, pouring liquids, manipulating cloth and cardboard, etc.), and fast execution to provide for high throughput. We illustrate the robotic platform used in our experiments in Figure~\ref{fig:robot}. We give details on the tasks and baselines below, followed by quantitative experiments.

\subsection{Evaluation Tasks}
Our quantitative evaluations and comparisons use three broad task categories each with individual task variants: laundry folding, coffee making, and box assembly. 
We summarize the tasks below, with illustrations in Figure~\ref{fig:tasks}:

\noindent \textbf{Laundry (t-shirts and shorts).} This is the standard laundry folding task in the $\pi_0$ paper~\citep{black2024pi_0}. This task entails retrieving either a T-shirt or shorts from a basket with variable initial conditions, flattening, folding. Success requires one clothing item to be folded and stacked in the top right corner of the table within 200 seconds.

\noindent \textbf{Laundry (diverse items).} The diverse laundry task requires folding a much larger variety of items, considering 11 item types, including towels, button-up shirts, sweaters, jeans, T-shirts, shorts, polos, skirts, long sleeve shirts, socks, and underwear. To obtain a low-variance metric in our experiments, we measure performance on one of the most challenging items -- the button-up shirt. However, the policy is trained on all items, and the accompanying videos show results for a variety of clothing. Success is defined as having the target item correctly folded and placed on a stack on the table within 500 seconds.

\noindent \textbf{Laundry (targeted failure removal).} The final version of the laundry folding task considers a much more structured setup for use in our ablation experiments, in which the task involves folding a single orange T-shirt from a fixed flattened initial condition. We place the highest emphasis on success, with a strict success criteria that requires the shirt to be folded correctly with the collar always facing up within 200 seconds. We found this task to be useful for assessing whether \MethodName{} can remove specific undesirable behaviors via RL (in this case, placing the collar facing down rather than up).

\noindent \textbf{Cafe (double shot espresso).} We evaluate our policies on the challenging long-horizon task of making coffee with a commercial espresso machine. While our cafe policy can make many drinks (lattes, iced Americanos, espresso, etc), and even clean the espresso machine with a towel, for the purposes of our quantitative experiments we focus on the double espresso shot task. This entails picking up the portafilter, placing it on the grinder and grinding beans into it, tamping the ground coffee beans, locking the portafilter into the espresso machine, bringing over the cup, extracting the full shot of espresso, then serving. Success is measured as completing all steps within 200 seconds without critical mistakes (such as dropping the portafilter or spilling the coffee).

\noindent \textbf{Box assembly.} We evaluate our policy on the problem of assembling packaging boxes in a real-world factory deployment scenario. Box assembly involves folding a cardboard box starting from a flattened cardboard sheet, attaching a label onto it and placing the box in the appropriate spot in a crate. For the purposes of the quantitative experiments, we focus on all portions of the task and count overall success as going from a flattened to an assembled and stacked box in under 600 seconds. 

\subsection{Comparisons and Ablations}

\begin{figure*}[ht]
    \centering
    \begin{subfigure}{0.26\textwidth}
        \centering
        \includegraphics[width=\linewidth]{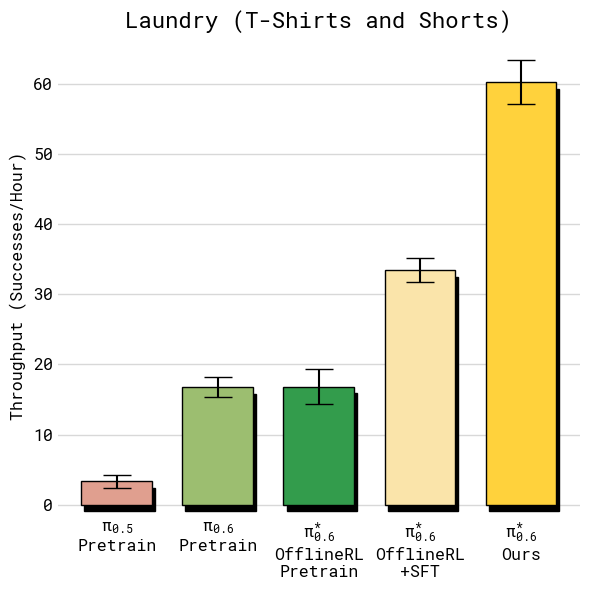}
    \end{subfigure}
    \begin{subfigure}{0.26\textwidth}
        \centering
        \includegraphics[width=\linewidth]{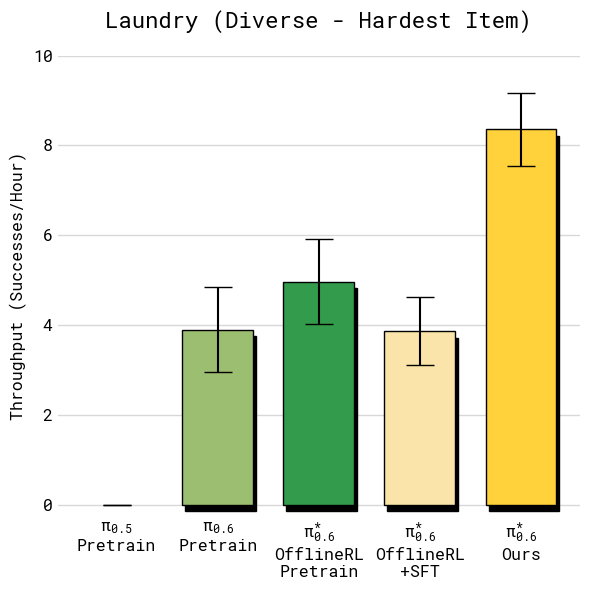}
    \end{subfigure}
    \begin{subfigure}{0.217\textwidth}
        \centering
        \includegraphics[width=\linewidth]{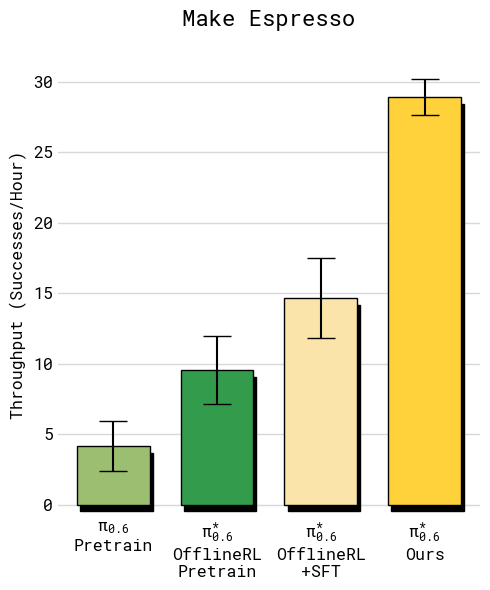}
    \end{subfigure}
     \begin{subfigure}{0.217\textwidth}
        \centering
        \includegraphics[width=\linewidth]{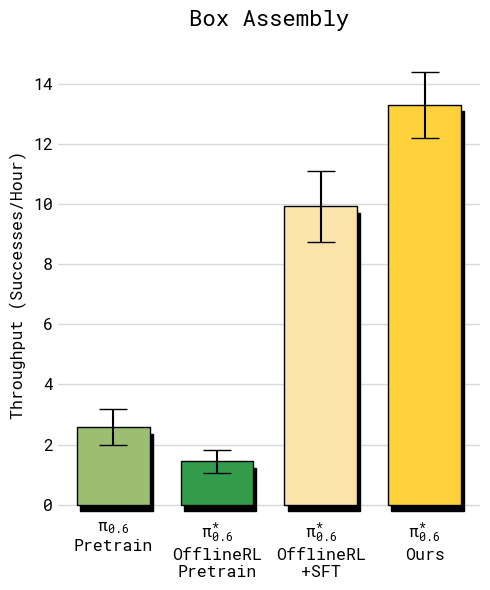}
    \end{subfigure}
    \caption{\textbf{Throughput.} We show the number of successfully completed tasks \emph{per hour} for laundry (simple and diverse), espresso making, and box assembly. Error bars show standard error. This metric measures both success and speed. In all cases, \MethodName{} applied to \ModelSymbol{} (Ours) leads to substantial improvements in throughput. \MethodName{} has the highest impact on throughput for diverse laundry and espresso tasks, more than doubling successful completions per hour.}
    \label{fig:main_throughput}
\end{figure*}

\begin{figure*}[ht]
    \centering
    \begin{subfigure}{0.225\textwidth}
        \centering
        \includegraphics[width=\linewidth]{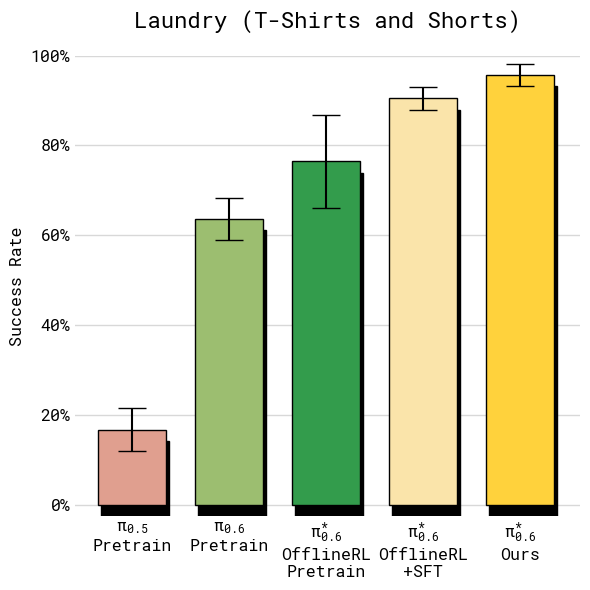}
    \end{subfigure}
    \begin{subfigure}{0.225\textwidth}
        \centering
        \includegraphics[width=\linewidth]{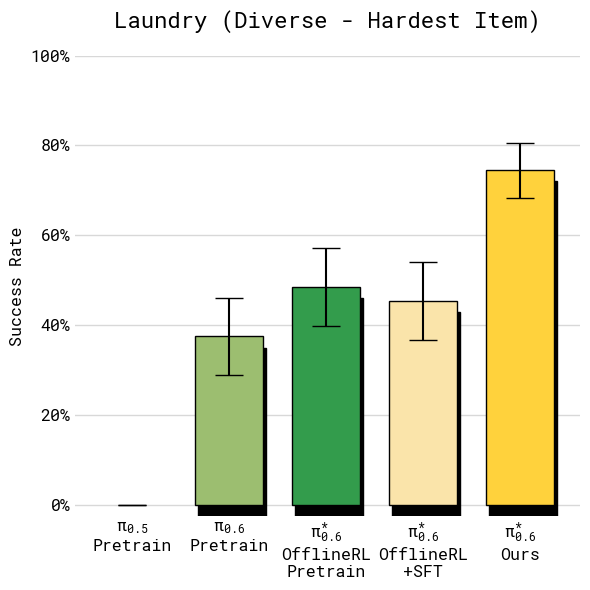}
    \end{subfigure}
    \begin{subfigure}{0.186\textwidth}
        \centering
        \includegraphics[width=\linewidth]{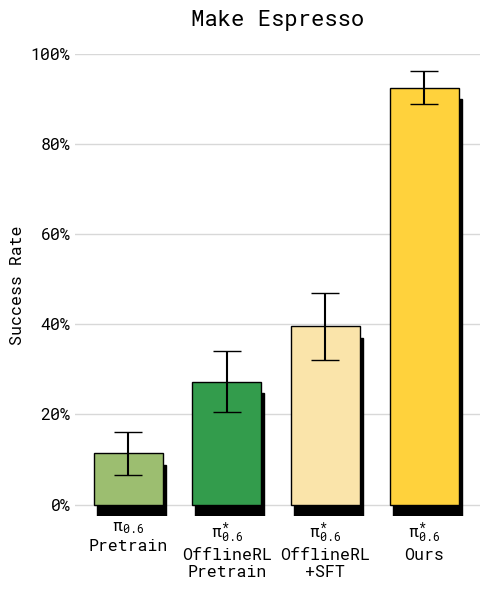}
    \end{subfigure}
    \begin{subfigure}{0.32\textwidth}
        \centering
        \includegraphics[width=\linewidth]{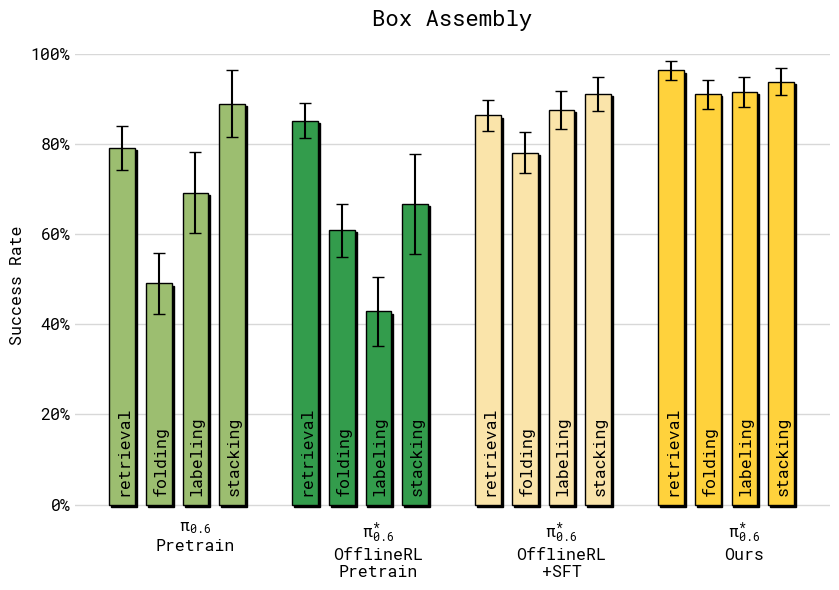}
    \end{subfigure}
    \caption{\textbf{Success rates.} We show the absolute success rates with standard error. Each stage of \MethodName{} improves performance across the tasks, with the challenging diverse laundry and espresso tasks seeing the largest gains success rate, corresponding to more than 2$\times$ reduction in failure rates. For the box assembly task we show the success rate for the different subtasks. \MethodName{} leads to the most consistent (and highest) success across all subtasks.
    }
    \label{fig:main_success}
\end{figure*}

We compare \MethodName{} to several baselines:

\noindent \textbf{Pre-trained \Pizf{}}~\citep{black2025pi05}. This baseline does not use RL and does not leverage \MethodName{}.

\noindent \textbf{Pre-trained \Pizs{}}~\citep{pi06model}. It does not include the advantage indicator $I_t$, and is pre-trained with supervised learning.

\noindent \textbf{RL pre-trained \ModelSymbol{}}. It is pre-trained with RL alongside its value function, and includes an advantage indicator $I_t$ as described in Section~\ref{sec:training}.

\noindent \textbf{\ModelSymbol{} offline RL + SFT}. This model is trained by finetuning the base \ModelSymbol{} pre-trained checkpoint with demonstration data for the target task. We refer to this finetuning as ``SFT" because the advantage values are fixed to $\text{True}$ for all demonstrations. We find that this combination of the offline RL pre-trained \ModelSymbol{} model with high-quality SFT outperforms standard SFT (without offline RL pre-training), and provides a good starting point for RL with on-robot data.

\noindent \textbf{\ModelSymbol{} (ours)}. This is the final model trained with \MethodName{} on the target task, including both autonomous rollouts and expert corrections. By default we evaluate with $\beta = 1$. In some experiments we also consider inference with CFG, which corresponds to $\beta > 1$. 

\noindent We also consider two alternative policy extraction methods in the literature as comparisons for our advantage-conditioned approach, both of which use the same on-robot data as \MethodName{} but a different policy learning method:

\noindent \textbf{AWR}. Starting from the same pre-trained model \Pizs{} (without advantage conditioning) we fine-tune using advantage weighted regression \citep{peng2019advantage}, based on advantages extracted from our value-function.

\noindent \textbf{PPO}. We implement a variant of DPPO/FPO \citep{Ren2025DPPO,mcallister2025fpo} in which we calculate likelihoods based on the single step diffusion objective and use an alternative definition of the PPO constraint following SPO \citep{xie2025simplepolicyoptimization} (see Appendix \ref{sect:ppo} for details).

\subsection{Quantitative results}

We use two metrics in our evaluation: throughput and success rate. Throughput measures the number of successful task executions per hour, thus capturing both speed and success rate into one practically relevant quantity. Success rate measures the proportion of episodes that succeed, and is derived from human-provided annotations. Raters are asked to judge the episode with respect to multiple quality metrics, and we aggregate these quality indicators into a success label.

\subsubsection{How much does \MethodName{} improve the policy?}

To answer this question, we present the main quantitative results in Figures \ref{fig:main_throughput} and \ref{fig:main_success}. Across all tasks, the final \ModelSymbol{} significantly improves over the base (supervised) \Pizs{} model, the RL pre-trained \ModelSymbol{} model, and the \textbf{offline RL + SFT} \ModelSymbol{} model. Throughput more than doubles on the diverse laundry folding and espresso tasks from including on-robot data (the improvement from \textbf{offline RL + SFT} to the final \ModelSymbol{} model), and the rate of failure reduces by about a factor of two. On the easier laundry task (t-shirts and shorts), the success rate is already close to the maximum after the SFT phase, but throughput still increases by a significant margin with the final model.

On all of the tasks except diverse laundry, the success rate of the final \ModelSymbol{} model is in the 90\%+ range. This makes it feasible to use in practical settings, such as making espresso drinks at the office or assembling boxes in a factory, as shown in the accompanying videos. For the box assembly task, Figure~\ref{fig:main_success} (right) contains a breakdown of the task success over its four stages: picking up a box sheet, building the box, labeling the box, and placing it at an available spot in a crate. \ModelSymbol{} attains higher success rates for all of the stages compared to the other models. The majority of failures on these stages happen because the policy runs out of time. The accompanying videos present time lapses where each of the tasks is run for multiple hours.

\begin{figure}[t]
    \centering
    \begin{subfigure}{0.22\textwidth}
        \centering
        \includegraphics[width=\linewidth]{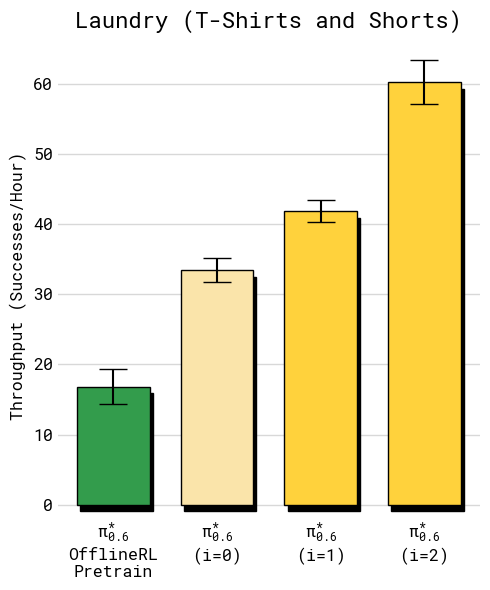}
    \end{subfigure}
    \begin{subfigure}{0.22\textwidth}
        \centering
        \includegraphics[width=\linewidth]{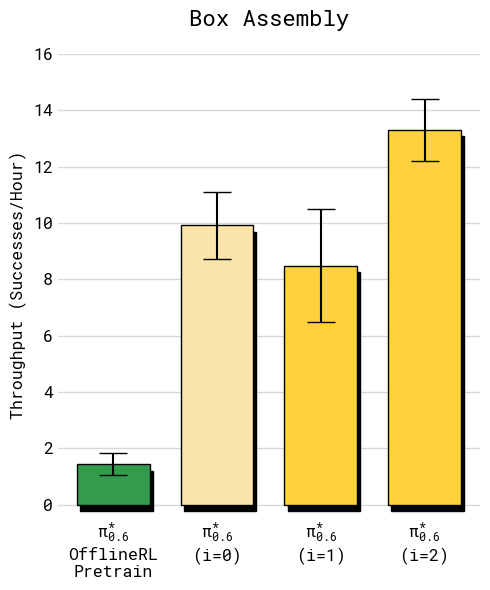}
    \end{subfigure}
    \caption{\textbf{Improvement in throughput over multiple iterations.} Both tasks improve significantly in throughput as we take more iterations of \MethodName{}, with box assembling first dropping and then improving significantly.}
    \label{fig:iter_throughput}
\end{figure}
\subsubsection{How much does \MethodName{} improve \ModelSymbol{} over multiple iterations?}

We next elucidate how training with \MethodName{} improves policies through multiple iterations of data collection and training. We study the T-shirt and shorts folding task and the box assembly task.
For the T-shirt folding task, only data collected with autonomous evaluation (without human corrections) is used to perform policy improvement over two iterations, in order to evaluate how well our method can improve the policy via RL alone. We collect 300 trajectories on four robots in each iteration. Box assembly uses both autonomous trials and trials with expert teleoperator interventions, with 600 autonomous trials and 360 trials with interventions in each iteration.

\begin{figure}[t]
    \centering
    \begin{subfigure}{0.175\textwidth}
        \centering
        \includegraphics[width=\linewidth]{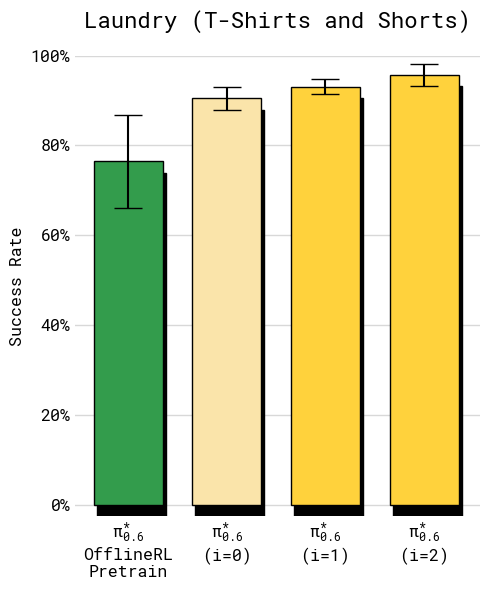}
    \end{subfigure}
    \begin{subfigure}{0.30\textwidth}
        \centering
        \includegraphics[width=\linewidth]{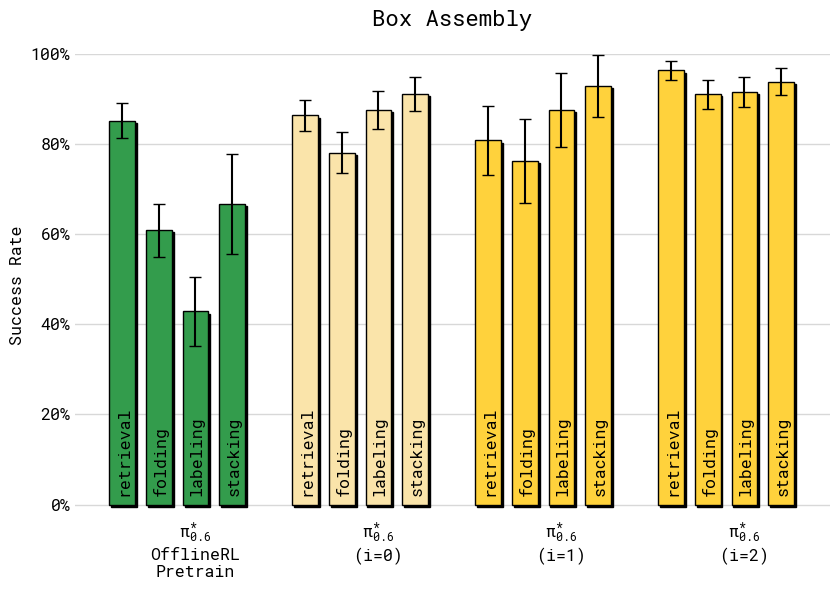}
    \end{subfigure}
    \caption{\textbf{Improvement in success rate over multiple iterations.} The laundry task quickly reaches the maximum success rate (but continues to improve in throughput as shown in Figure~\ref{fig:iter_throughput}, while box assembly continues to improve.}
    \label{fig:iter_success}
\end{figure}
We plot the throughput over iterations in Figure \ref{fig:iter_throughput}, comparing two iterations of \MethodName{}, denoted by $i = 1$, $i = 2$ respectively. The final iteration, labeled (Ours), corresponds to the overall best result for these tasks presented in the previous section. We also compare the initial data collection policy, which uses the offline RL pre-trained \ModelSymbol{} model with SFT finetuning. For both tasks, \ModelSymbol{} improves over the two iterations. In the laundry task we can see steady improvement yielding an overall $50 \%$ improvement in throughput. For the long-horizon box assembly task, more data is needed to yield a significant improvement, but after the second iteration we see a 2$\times$ improvement in throughput.
\begin{figure}[t]
    \centering
    \begin{subfigure}{0.24\textwidth}
        \centering
        \includegraphics[width=\linewidth]{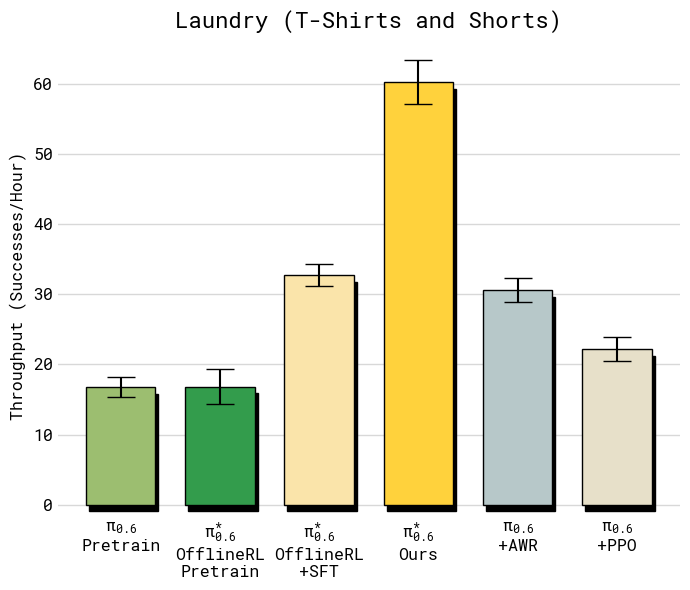}
    \end{subfigure}
     \begin{subfigure}{0.24\textwidth}
        \centering
        \includegraphics[width=\linewidth]{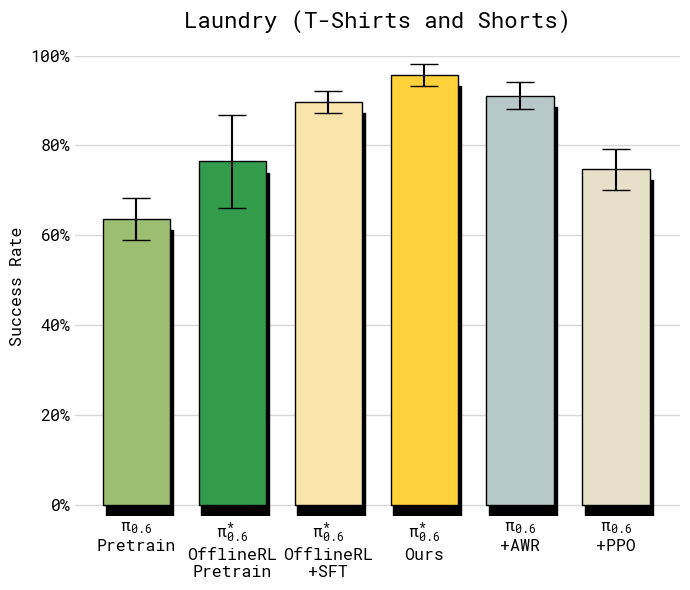}
    \end{subfigure}
    \caption{\textbf{Comparison of different policy extraction methods.} \MethodName{} applied to \ModelSymbol{} achieves by far the highest throughput for the laundry task compared to AWR and PPO.}
    \label{fig:baseline_comp}
\end{figure}

We also show the success rate over the iterations in Figure~\ref{fig:iter_success}. For the laundry task, the first iteration already raises the success rate to over 90\%, while the second iteration mainly improves throughput. For the box assembly task, we see clear improvements in the success rate over both iterations. While there are still some failures (especially when placing the box on the stack at the end), the final policy achieves a success rate of about 90\% both for folding the box and labeling it in the allocated time limit of $600$ seconds.

\subsubsection{How does the advantage-conditioned policy extraction method in \MethodName{} compare to other methods?}
We compare our advantage conditioned policy extraction method from Section~\ref{sect:adv_cond} to other methods in the literature: AWR and PPO. We use the T-shirts and Shorts task for this comparison. To ensure a controlled comparison, we use the same data for these comparisons that was used to train our final model. This provides a slight advantage to the baselines, since they have access to better data that was collected while running \MethodName{}. The results are shown in Figure~\ref{fig:baseline_comp}. While both AWR and PPO can attain reasonable results, they both fall far short of our method, and struggle to improve over the offline RL + SFT \ModelSymbol{} model. For PPO, we had to use a small trust-region constraint ($\eta = 0.01$) to stabilize training in this off-policy setting, and while this makes training stable, the method does not achieve good performance. AWR can achieve a reasonable success rate, but leads to much slower polies with lower throughput.

\begin{figure}[t]
    \centering
    \begin{subfigure}{0.24\textwidth}
        \centering
        \includegraphics[width=\linewidth]{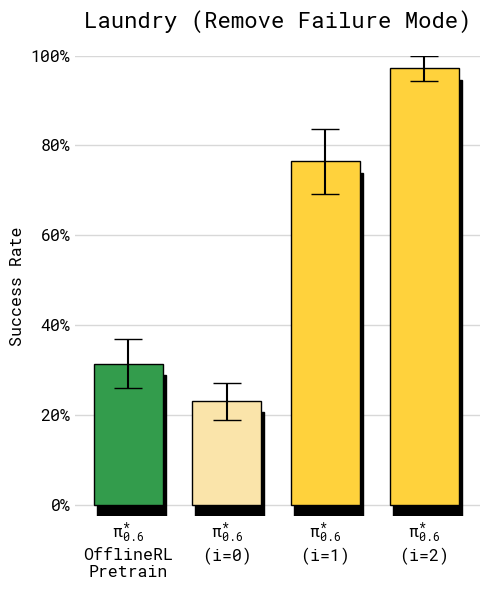}
    \end{subfigure}
    \begin{subfigure}{0.24\textwidth}
        \centering
        \includegraphics[width=\linewidth]{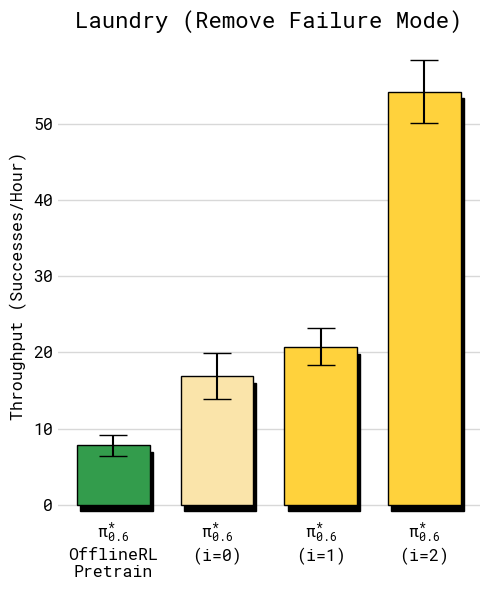}
    \end{subfigure}
    \caption{\textbf{Failure mode removal.} Here we apply \MethodName{} on a variant of the laundry task with one item but a very strict success criteria. \MethodName{} is particularly effective at removing failure modes that would be considered non successful under the strict criteria. Therefore, our method can also be used to alter a policy's behavior with relatively little data effectively.}
    \label{fig:failure_removal}
\end{figure}

\subsubsection{Can \MethodName{} significantly alter policy behavior with relatively little data and remove a failure mode?}
While the preceding experiments have focused on holistic end-to-end evaluations of policy performance, we can also zoom in on a specific failure mode to examine whether RL training with \MethodName{} can remove a specific mistake from the policy. To answer this question, we use a version of the laundry task with a strict success criterion, which requires the policy to fold a t-shirt with the collar centered and facing up. Each episode is initialized with a specific adversarial condition in which the shirt is placed flat on the table in such a way that the baseline offline RL + SFT policy often fails to fold it correctly. As shown in Figure~\ref{fig:failure_removal}, applying \MethodName{} in this setting for two iterations (collecting $600$ trajectories in each iteration) results in a policy that succeeds $97 \%$ of the time, and with high speed. Thus we conclude that \MethodName{} can be effective at removing specific failure modes, even when learning entirely via RL without any intervention data or additional demonstrations.

\section{Discussion and Future Work}

Training policies that can achieve the same robustness, speed, and fluency on real-world tasks as people presents a major challenge in robotic learning. In this paper, we discussed how learning from experience, through a combination of DAgger-style coaching and RL, can begin to address this challenge. We describe \MethodName{}, a method for training VLAs with autonomous trials, reward feedback, and human interventions, and present results for a model trained with \MethodName{}, \ModelSymbol{}, on a set of realistic tasks: making espresso drinks, folding diverse laundry, and assembling boxes. At the core of \MethodName{} is an RL method that is well-suited for scalable training of VLA policies, using advantage conditioning for policy extraction with value functions. The data for this RL method is collected with a combination of autonomous rollouts and human interventions, correcting mistakes with interventions while finetuning the details of the behavior on autonomous data. Our experiments show that \MethodName{} can improve both the success rate and throughput of the VLA, more than doubling the throughput on some of the harder tasks, and decreasing the number of failures by roughly 2$\times$.

There are several directions for improvement with \MethodName{}. First, our system is not fully autonomous: it relies on human labeling and effort for reward feedback, interventions, and episode resets. A number of prior works have explored ways to automate these components~\citep{zhu2020ingredients,sharma2021autonomous}, and VLAs offer new ways to provide for more automated data collection, for example by using high-level policies~\citep{shi2025hi} to reason through resetting the scene. Second, our system is relatively na\"{i}ve in how it approaches exploration: exploration is largely greedy, relying on stochasticity in the policy and human interventions to explore new solutions. This is reasonable when the initial imitation learning policy already takes reasonable actions, but there is plenty of room for improvement with more sophisticated exploration methods. Lastly, \MethodName{} performs iterated ``offline'' updates (i.e., it collects a batch of data, retrains the model, and repeats), rather than running a fully online RL loop where the policy and value function are updated in real time as data is collected. We make this decision out of convenience, but extending our approach into a fully concurrent online RL framework is a promising direction for future work.

More broadly, training VLAs with RL is perhaps the most direct path to get to performance levels that are adequate for real-world use cases. RL with VLAs presents a number of challenges, from the difficulty of large-scale RL training of high capacity models to sample complexity, autonomy, and delayed feedback. While existing RL frameworks designed for smaller-scale systems or ``virtual'' domains such as LLMs can provide a good starting point, more research will be needed to make RL a practical tool for VLA training. We hope that our work represents a meaningful step in this direction.

\section*{Acknowledgements}

We thank our robot operators for data collection, evaluations, logistics, and video recording, and our technicians for robot maintenance and repair. See Appendix~\ref{app:contributions} for a full contributions statement.

\bibliographystyle{unsrtnat}
\bibliography{references}

\appendix

\section{}

\subsection{Contributions}
\label{app:contributions}

\noindent\textbf{Data collection and operations}. Michael Equi, Chelsea Finn, Lachy Groom, Hunter Hancock, Karol Hausman, Rowan Jen, Liyiming Ke, Marinda Lamb, Vishnu Mano, Suraj Nair, Charvi Sharma, Laura Smith, Will Stoeckle, Anna Walling, Blake Williams.

\noindent\textbf{Annotation and supplemental data}. Chelsea Finn, Catherine Glossop, Hunter Hancock, Brian Ichter, Rowan Jen, Liyiming Ke, Chandra Kuchi, Karl Pertsch, Laura Smith, Will Stoeckle, Quan Vuong, Anna Walling.

\noindent\textbf{Policy training and research}. Ashwin Balakrishna, Kevin Black, Danny Driess, Michael Equi, Yunhao Fang, Chelsea Finn, Catherine Glossop, Karol Hausman, Gashon Hussein, Brian Ichter, Liyiming Ke, Sergey Levine, Yao Lu, Suraj Nair, Karl Pertsch, Allen Z. Ren, Lucy Shi, Laura Smith, Jost Tobias Springenberg, Kyle Stachowicz, Alex Swerdlow, Marcel Torne, Quan Vuong, Lili Yu, Zhiyuan Zhou.

\noindent\textbf{Policy infrastructure}. Kevin Black, Karan Dhabalia, Danny Driess, Michael Equi, Liyiming Ke, Adrian Li-Bell, Suraj Nair, Allen Z. Ren, Laura Smith, Jost Tobias Springenberg, Kyle Stachowicz, Alex Swerdlow, Haohuan Wang, Ury Zhilinsky, Zhiyuan Zhou.

\noindent\textbf{Robot hardware}. Ali Amin, Raichelle Aniceto, Grace Connors, Adnan Esmail, Thomas Godden, Ivan Goryachev, Tim Jones, Ben Katz, Devin LeBlanc, Mohith Mothukuri, Sukwon Yoo.

\noindent\textbf{Robot infrastructure}. Ken Conley, James Darpinian, Jared DiCarlo, Karol Hausman, Szymon Jakubczak, James Tanner.

\noindent\textbf{Writing and illustration}. Kevin Black, Danny Driess, Michael Equi, Chelsea Finn, Hunter Hancock, Karol Hausman, Brian Ichter, Liyiming Ke, Sergey Levine, Suraj Nair, Allen Z. Ren, Laura Smith, Jost Tobias Springenberg, Zhiyuan Zhou

\subsection{Additional Value Function Visualization}
\label{app:value}

\begin{figure}
    \centering
    \vspace{1em}
    \includegraphics[width=0.99\linewidth]{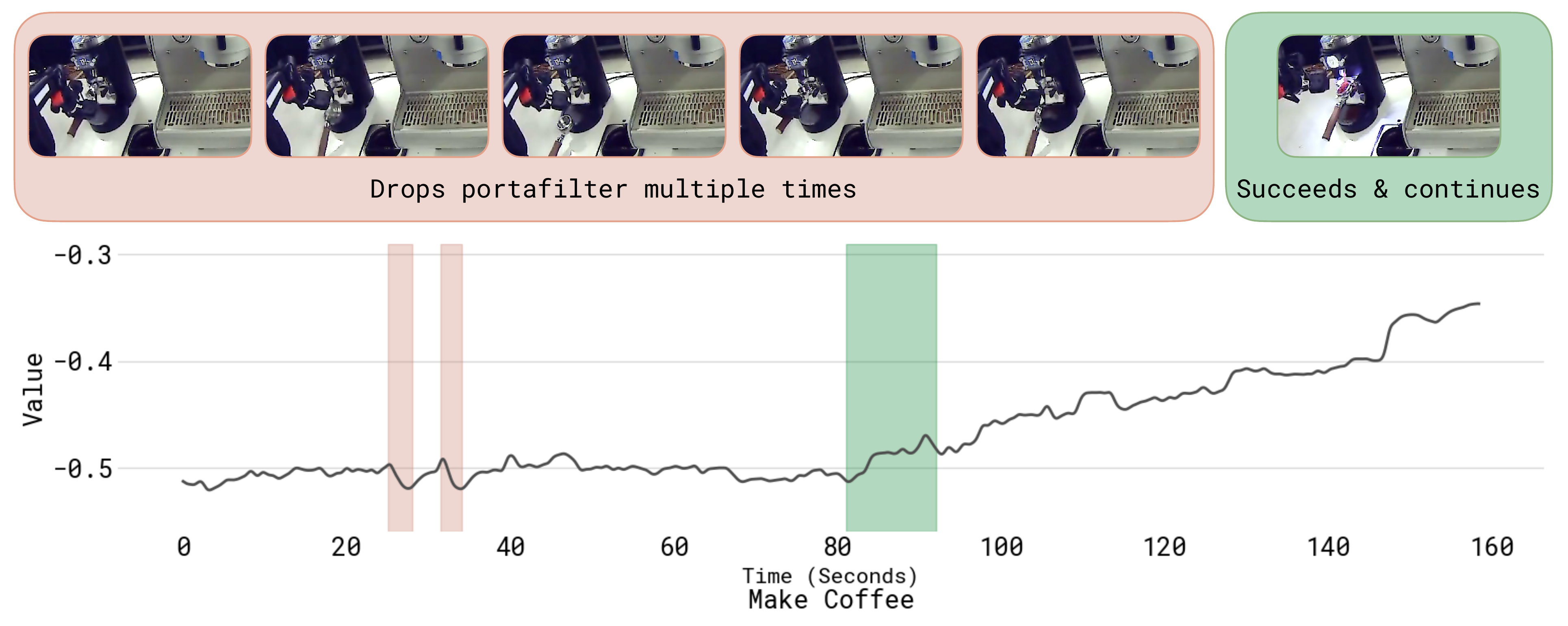}
    \vspace{1em}
    \includegraphics[width=0.99\linewidth]{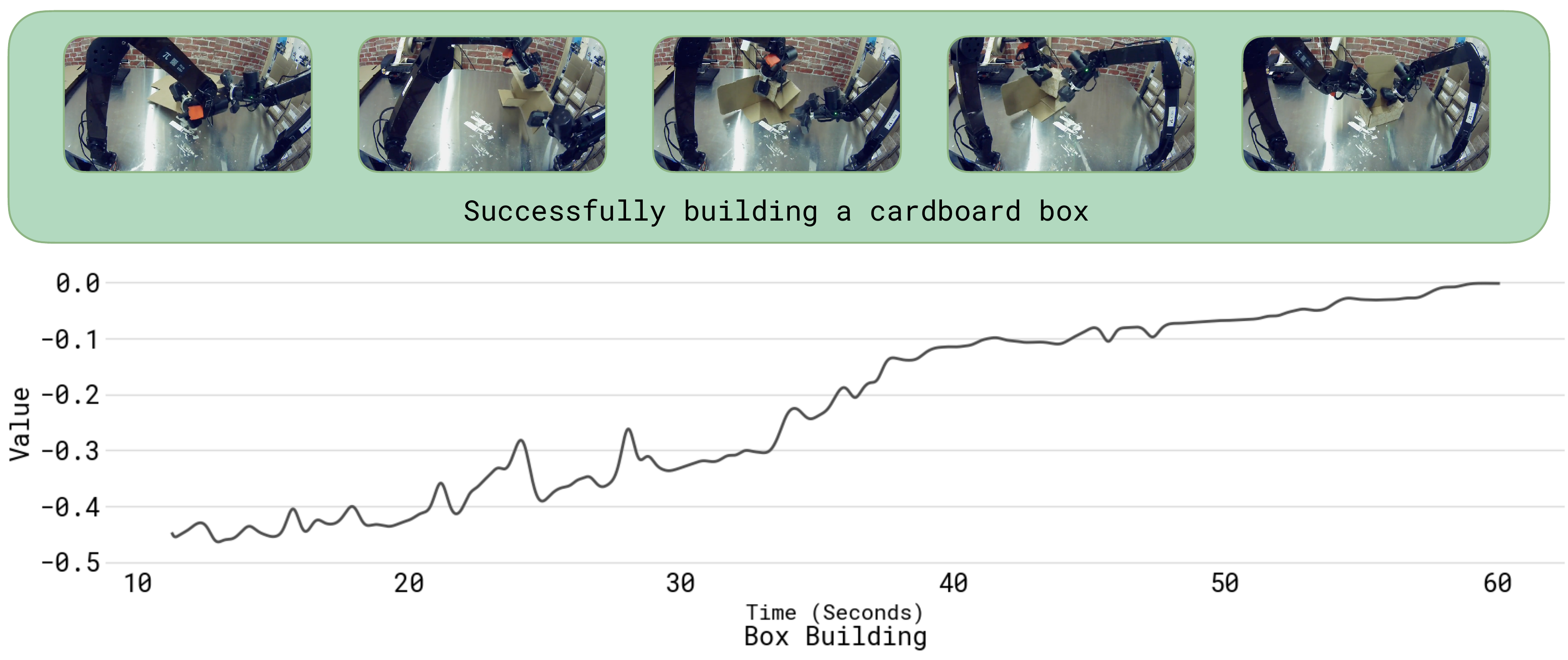}
    \vspace{1em}
    \includegraphics[width=0.99\linewidth]{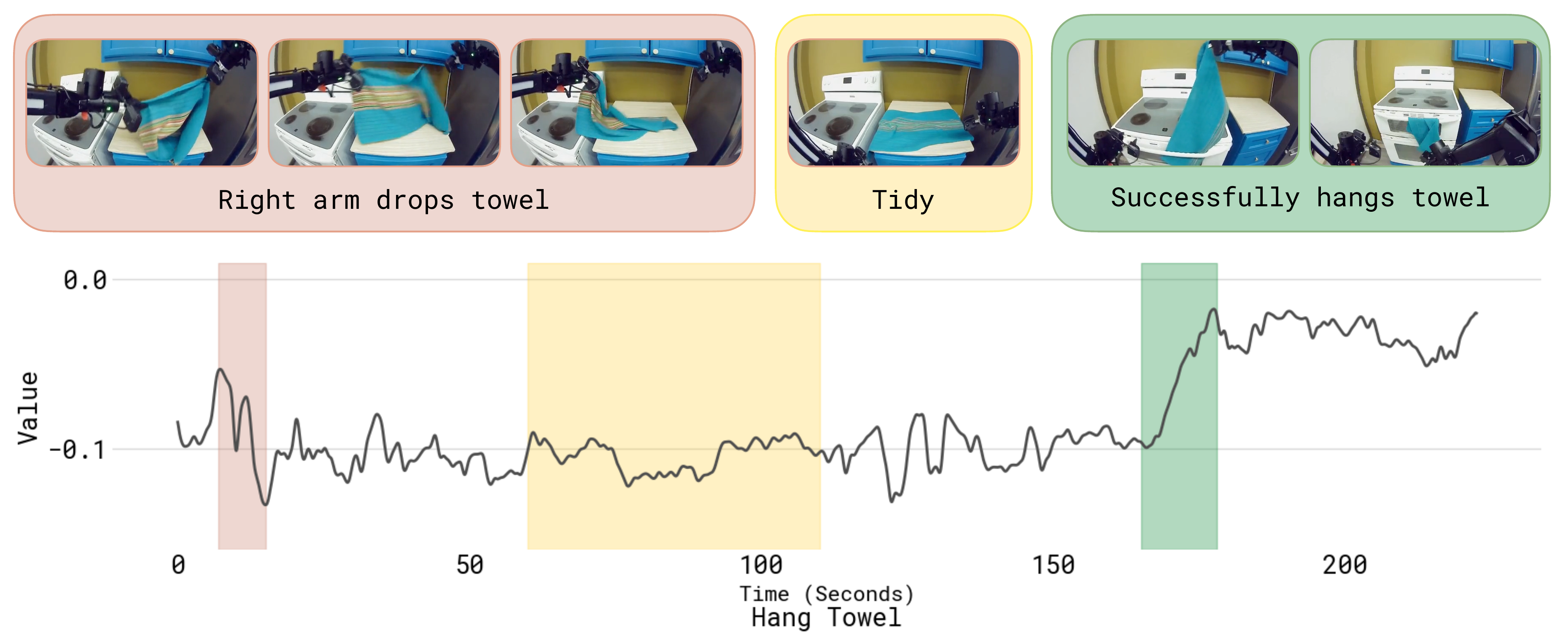}
    \vspace{1em}
    \includegraphics[width=0.99\linewidth]{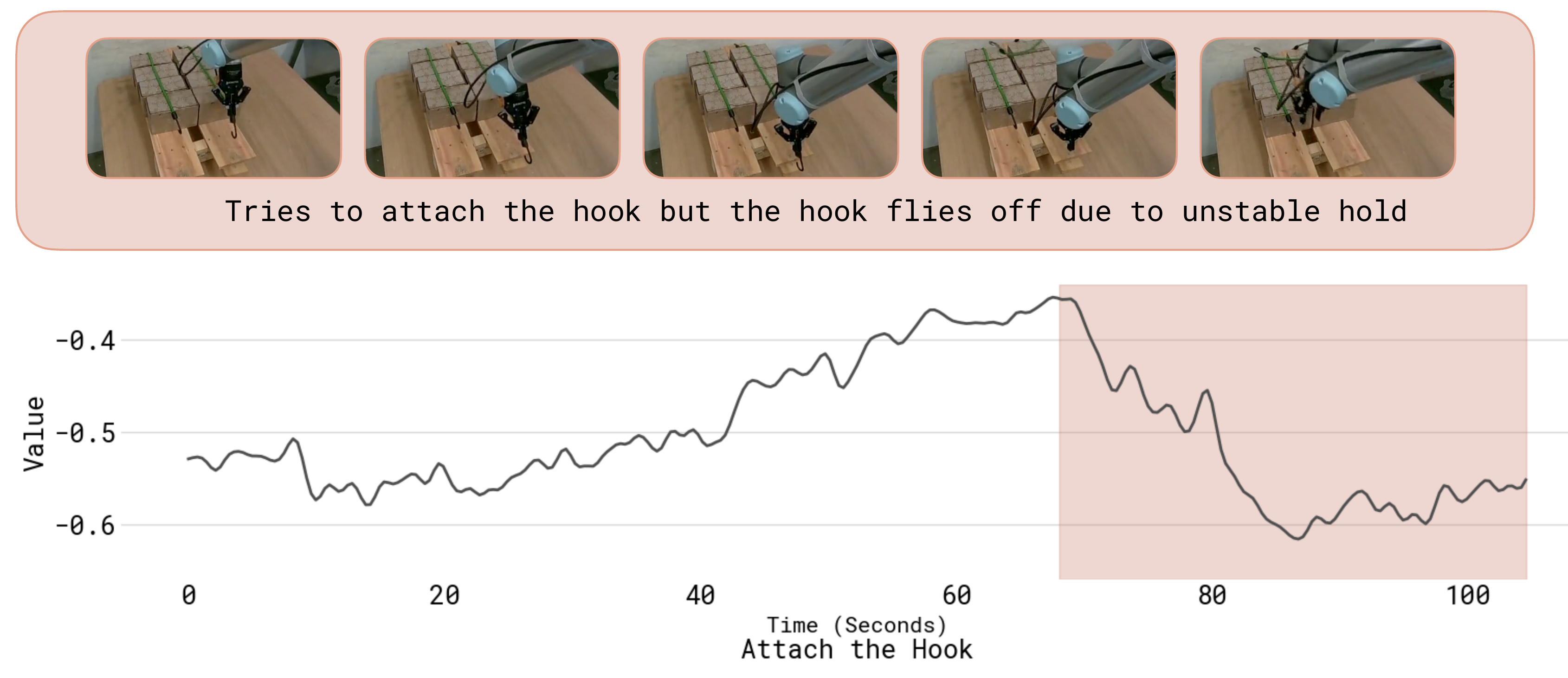}
    \caption{\textbf{Additional visualization of value function on five different tasks.} Red parts highlight places where value drops, green parts highlight places where value increases, and yellow parts highlight oscillating value regions. Images show the corresponding frames and descriptions of the episode.}
    \label{fig:app-vf-viz}
\end{figure}

Figure~\ref{fig:app-vf-viz} shows additional visualizations of our trained value function on five different tasks, including tasks on which we evaluate our policies (espresso making, box assembly) and also broader tasks (hang towel, attach hook). The parts with the most prominent changes are highlighted: red corresponds to where value function drops, green corresponds to where value function increases, and yellow corresponds to oscillating values. Images show the corresponding frames and description of the episode.
\label{app:value-function-viz}

\subsection{Computing the log-likelihood for policy improvement}
\label{sect:lower_bound_loss}
To derive the log-likelihood from Equation~\eqref{eq:cotraining} we can first observe that we can decompose the full model likelihood into autoregressive and diffusion terms
\begin{equation}
\begin{aligned}
&\pi_\theta(\ba_{t:t+H}, \ba^{\lang}_{t:t+H}, \rawtext \vert I_t, \bo_t, \lang) = \\ &\pi_\theta(\ba_{t:t+H} \vert I_t, \bo_t, \lang, \rawtext) \pi_\theta(\ba^{\lang}_{t:t+H} \vert I_t, \bo_t, \lang, \rawtext) \pi_\theta(\rawtext \vert I_t, \bo_t, \lang),
\end{aligned}
\end{equation}
where the first term is modeled with flow matching, the second term is the autoregressive likelihood of the discretized actions $\ba^{\lang}_{t:t+H}$, and the third term corresponds to the autoregressive text likelihood. The autoregressive likelihoods can be estimated in the usual way, using the cross-entropy loss evaluated on ground truth tokens. For the continuous likelihood over $\ba_{t:t+H}$, a closed form likelihood is not available~\citep{lipman2022flow}. We can, however follow prior work \citep{mcallister2025fpo}, and consider the one-step diffusion process as a Gaussian distribution with likelihood
\begin{equation}
\begin{aligned}
    \log &\pi_\theta(\ba_{t:t+H} \vert \ba^{\eta, \omega}_{1:H}, I_t, \bo_t, \lang, \hat{\ell}) = \\
    &\log \mathcal{N}\Big(\omega - f_\theta(\ba^{\eta, \omega}_{1:H}, I_t, \bo_t, \lang, \hat{\ell}),  \mathbf{I} \Big), 
\end{aligned}
\end{equation}
with  
$\ba_{t:t+H}^{\eta, \omega} = \eta \ba_{t:t+H} + (1-\eta)\omega$ and $\omega = \mathcal{N}(0, \bI)$. From this we can form an evidence lower bound to the likelihood following \citep{KingaDiffELBO, mcallister2025fpo} (effectively marginalizing over $\eta$ and $\omega$) which yields
\begin{equation}
\begin{aligned}
\log &\pi_\theta(\ba_{t:t+H} \vert I_t, \bo_t, \lang, \hat{\ell}) \geq \\
&\frac{1}{2} \mathbb{E}_{\eta, \omega} \Big[ - w(\eta)  \left\|\omega - \ba_{1:H} - f_\theta(\ba^{\eta, \omega}_{1:H}, I_t, \bo_t, \lang, \hat{\ell})\right\|^2  \Big] + c,
\end{aligned}
\end{equation}
where $w(\eta) = e^{-\eta/2}$ is a noise dependent weighting term, and c is a constant independent of $f_\theta$. For the derivation, see \citep{KingaDiffELBO}, which also derives the relationship between flow matching and diffusion in Appendix D.3 for this choice of weighting term.
Finally putting the lower bound together with the autoregressive likelihood for the discretized action part of the text output $\rawtext$, and subsuming the weighting terms in $\alpha$, gives  
\begin{equation}
\begin{aligned}
    \log \pi_\theta(&\ba_{t:t+H}, \ba^{\lang}_{t:t+H} \vert I_t, \bo_t, \lang, \hat{\ell}) \geq \\  \mathbb{E}_{\eta, \omega} \Big[&  \log p_\theta(\ba^{\lang}_{t:t+H} | I_t, \bo_t, \lang, \hat{\ell})
     \\ &- \alpha_\eta \left\|\omega - \ba_{1:H} - f_\theta(\ba^{\eta, \omega}_{1:H}, I_t, \bo_t, \lang, \hat{\ell})\right\|^2 \Big],
\end{aligned}
\label{eq:likelihood_bound_appendix}
\end{equation}
which is the bound given in the main part of the paper.

\subsection{PPO implementation}
\label{sect:ppo}
We implement a variant of PPO \citep{schulman2017proximal} related to DPPO and FPO \citep{Ren2025DPPO,mcallister2025fpo} and use it as an additional baseline. To allow for training both the autoregressive part of the model as well as the diffusion based action expert in a compute effective manner we calculate likelihoods based on the single step diffusion objective alone. 

In particular, we use a likelihood bound analogous to Eq. \eqref{eq:likelihood_bound_appendix} (previous section) but without the improvement indicator. Decomposing into autoregressive and flow-matching terms this can be written as
\begin{equation}
\begin{aligned}
    \log \pi_\theta(&\ba_{t:t+H}, \ba^{\lang}_{t:t+H} \vert \bo_t, \lang, \hat{\ell}) \geq \\  \mathbb{E}_{\eta, \omega} \Big[&  \log p_\theta(\ba^{\lang}_{t:t+H} | \bo_t, \lang, \hat{\ell})
     \\ &- \alpha_\eta \left\|\omega - \ba_{1:H} - f_\theta(\ba^{\eta, \omega}_{1:H}, \bo_t, \lang, \hat{\ell})\right\|^2 \Big],
\end{aligned}
\end{equation}
which is analogous to the diffusion likelihood bound used in FPO \citep{mcallister2025fpo}. And we combine it with a PPO style loss separated into diffusion and autoregressive terms. In preliminary experiments we found that for our setting it was difficult to enforce a trust region constraint on the action expert (which models actions with an unbounded diffusion head) when using the standard PPO clipping objective. Presumably, this is partially due to the ``offline'' nature of our algorithm setting, where we cannot afford to collect new data from real robots every few gradient steps.
To stabilize training we found using an alternative definition of the PPO constraint following SPO \citep{xie2025simplepolicyoptimization} to be effective. The resulting loss is given as:
\begin{equation}
\begin{aligned}
&\mathcal{L}_{SPO+CoVLA}(\theta) =  \\  
&\Bigg\lbrace \frac{\pi_\theta( a_{\rawtext} \in \rawtext | \bo_t, \lang)}{\piref( a_{\rawtext} \in \rawtext | \bo_t, \lang)} A^{\piref}(o_t, a_t, \lang) \\
&- \frac{|A^{\piref}(o_t, a_t, \lang)|}{2\epsilon_{\text{ar}}} \Bigg[\frac{\pi_\theta( a_{\rawtext} \in \rawtext | \bo_t, \lang)}{\piref( a_{\rawtext} \in \rawtext | \bo_t, \lang)} - 1 \Bigg] \Bigg\rbrace \\
+ \alpha &\Bigg\lbrace \frac{\pi_\theta(\ba_{t:t+H} \vert \bo_t, \lang)}{\piref(\ba_{t:t+H} \vert \bo_t, \lang)} A^{\piref}(o_t, a_t, \lang) \\
&- \frac{|A^{\piref}(o_t, a_t, \lang)|}{2\epsilon_{\text{flow}}} \Bigg[\frac{\pi_\theta(\ba_{t:t+H} \vert \bo_t, \lang)}{\piref(\ba_{t:t+H} \vert \bo_t, \lang)}  - 1 \Bigg] \Bigg\rbrace,
\end{aligned}
\end{equation}
where $\alpha$ is a trade-off parameter and $\epsilon_{\text{ar}}$, $\epsilon_{\text{flow}}$ are trust-region parameters for autoregressive and flow-matching model parts respectively.
We use this variant to perform training on eval data starting from the \Pizs{}{} checkpoint.

\subsection{Using CFG for test-time policy improvement with $\beta > 1$}
\label{sect:cfg}
After training we can choose to further sharpen the policy used for evaluation by setting $\beta > 1$ in Eq. \eqref{eq:improved_policy}. As shown in prior work \citep{Frans2025DiffusionGuidance} we can recover this sharpened policy without additional training since it is implicitly defined by the learned policies 
$\pi_\theta(\ba_{t:t+H} \vert I_t, \bo_t, \lang)$ and $\pi_\theta(\ba_{t:t+H} \vert \bo_t, \lang)$.
Specifically, after training we can form the approximation
\begin{equation}
      \hat{\pi}(\ba_{t:t+H} \vert \bo_t, \lang) \propto \piref(\ba_{t:t+H} \vert \bo_t, \lang) \left( \frac{\piref(\ba_{t:t+H} \vert I_t, \bo_t, \lang)}{\piref(\ba_{t:t+H} \vert \bo_t, \lang)} \right)^\beta.
\end{equation}
One can now realize that the diffusion model effectively learns the gradient of the likelihoods, i.e. it represents $\nabla_\ba \log \pi_\theta(\ba_{t:t+H} \vert I_t, \bo_t, \lang)$ and $\nabla_\ba \log \pi_\theta(\ba_{t:t+H} \vert \bo_t, \lang)$ respectively. From this, following \citet{Frans2025DiffusionGuidance}, we can see that if we run flow-matching inference following the gradient
\begin{equation}
\begin{aligned}
    \nabla_\ba &\log \pi_\theta(\ba_{t:t+H} \vert \bo_t, \lang) + \\ 
    \beta &( \nabla_\ba \log \pi_\theta(\ba_{t:t+H} \vert I_t, \bo_t, \lang) - \nabla_\ba \log \pi_\theta(\ba_{t:t+H} \vert \bo_t, \lang)),
\end{aligned}
\end{equation}
we are effectively sampling from the desired attenuated distribution. We note that, as mentioned in the main paper, the parameter $\beta$ is loosely connected to the advantage threshold $\epsilon_\lang$ that we introduce during training (in the sense that both sharpen the distribution, one at inference and one at training  time). We find that sharpening the distribution after training with high settings for $\beta$ can lead to pushing the action distribution towards the boundaries of its learned support (which can lead to overly aggressive motions) and thus primarily rely on $\epsilon_\lang$ for obtaining a good conditioned policy directly after training and combine it with moderate settings (e.g. $\beta \in [1.5, 2.5]$) where useful.

\subsection{Additional algorithm details}
\label{app:alg_details}

We describe details for setting the task specific parameters used in Algorithm \ref{alg:summary}.

\textbf{Advantage Estimation:} During post-training, we estimate the advantage function using $A^\pi(\bo_t,\ba_t) = \sum_{t'=t}^{t+N-1} r_t' + V^\pi(\bo_{t+N}) - V^\pi(\bo_t)$, where $\bo_{t+N}$ is an observation sampled from $N$ steps ahead from the same trajectory. We use $N=50$ lookahead to calculate this advantage. During pre-training, we calculate the advantage estimate as $A^\pi(\bo_t,\ba_t) = \sum_{t'=0}^{T} r_t' - V^\pi(\bo_t)$, setting $N = T$ for each episode, which is a higher variance estimate of the advantage. We use this advantage calculation since it allows us to calculate the advantage values on-the-fly during pre-training using a single inference call to the value function. We find empirically that this advantage estimate works well when the policy is trained on large amounts of data from diverse tasks during pre-training.

\textbf{Advantage conditioning dropout:} During training, we randomly drop out the conditioning on the advantage indicator 30\% of the time. We employ this dropout so that we can directly sample directly from either the conditional or unconditional policy during inference time and use CFG for test-time policy improvement (see Section~\ref{sect:cfg} for details); and it effectively replaces the loss multiplier $\alpha$.

\textbf{Advantage threshold:} The per task advantage threshold $\epsilon_\lang$ is set as follows. During pre-training we select the threshold for each task such that approximately $30 \%$ of the demonstration data has positive advantage (as calculatedd on a random sample of 10k datapoints). During fine-tuning we generally set the threshold such that approximately $40 \%$ of the evaluation rollouts in each iteration have positive advantage. For the T-shirt and shorts laundry folding task (in which training on high-quality demonstration data yields slow policies but with high success rate) we increase the threshold such that only approximately $10 \%$ of the data has positive advantage.

\textbf{Dataset composition:} We use the dataset aggregation strategy described in Algorithm \ref{alg:summary} for all tasks. However each of our task has distinct nature: the episode lengths vary, the performances of Iteration 0 model on each task are different, and one task (Assemble Box) is performed offsite in a deployment scenario. Therefore, we have different amount of demonstration data to begin with and collect different amounts of experience data for iterative improvement. For laundry (T-shirt and shorts), we use autonomous evaluation data only without expert corrections. As we push model performance to closely resemble the expert data collector in terms of speed, it becomes hard to provide corrections. For this task, We collect 300 episodes across 4 robot stations for reporting eval performance. For the diverse laundry folding task we collect 450 evaluation episodes and 287 correction episodes. For the failure mode removal ablation we collect both autonomous and policy correction data. In total we collect $\sim 1000$ autonomous and $280 + 378$ correction episodes spread over $3$ robots. For box assembly we collect data in the deployment scenario directly, collecting $600$ demonstrations and $360$ correction episodes in each iteration, using $3$ robots in total. For cafe we perform a single iteration and collect $429$ correction episodes as well as $414$ autonomous episodes.

\end{document}